%% file: camera_ready.tex
\documentclass[10pt,twocolumn,letterpaper]{article}
\usepackage{cvpr}      
\usepackage{stfloats}
\usepackage{algorithmicx}
\usepackage[ruled]{algorithm2e}
\usepackage{multirow} 
\usepackage[accsupp]{axessibility} 
\usepackage{bbm}

\input{tex/preamble}

\input{tex/math_commands}

\input{tex/config}
\def\paperID{1126} 
\def\confName{CVPR}
\def\confYear{2024}

\usepackage{setspace}
\title{\textsc{PhyScene}: Physically Interactable 3D Scene Synthesis for Embodied AI}
\author{
Yandan Yang$^*$ \qquad Baoxiong Jia$^*$ \qquad Peiyuan Zhi \qquad Siyuan Huang \vspace{5pt}\\
State Key Laboratory of General Artificial Intelligence, \\ Beijing Institute for General Artificial Intelligence (BIGAI) \vspace{5pt}\\
\url{https://physcene.github.io}
\vspace{-15pt}
}

\begin{document}
\twocolumn[{
\renewcommand\twocolumn[1][]{#1}%
\maketitle
\begin{center}
    \centering
    \includegraphics[width=1\linewidth]{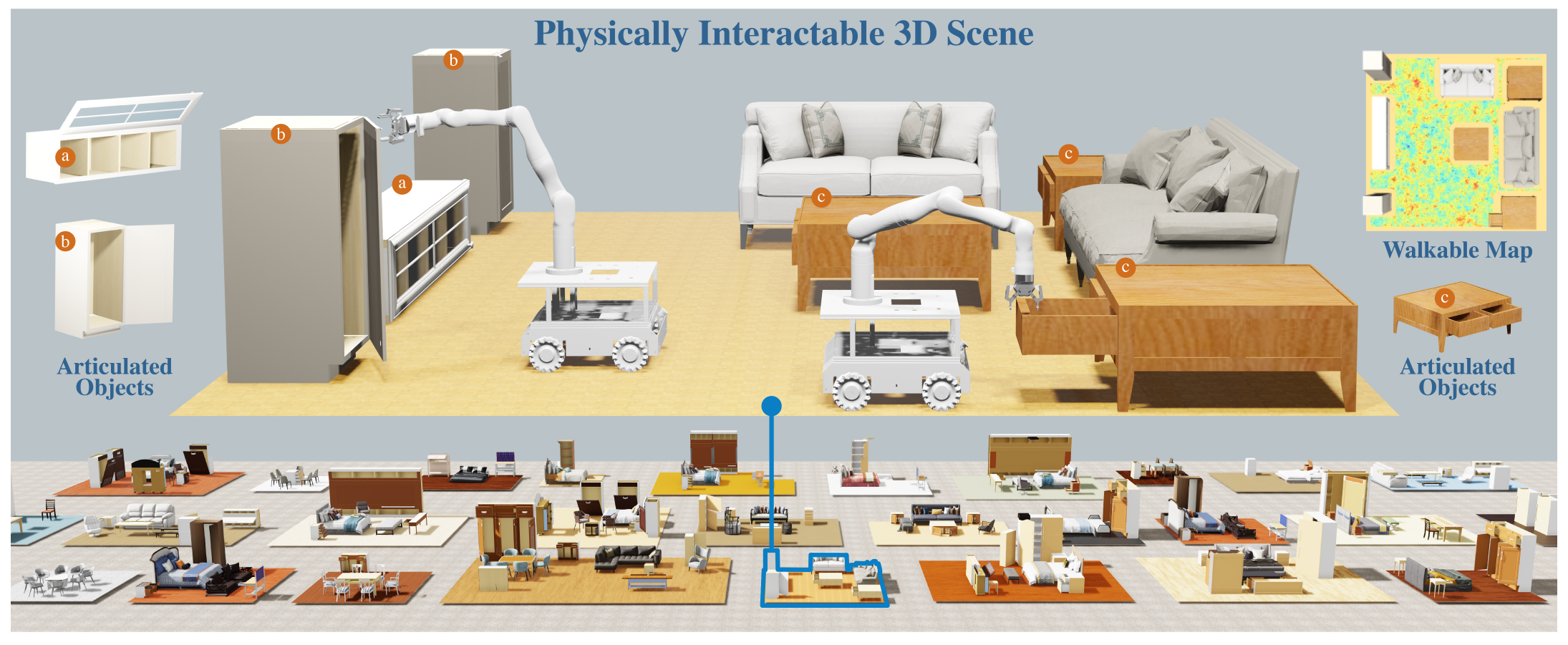}
    \captionsetup{type=figure}
    \vspace{-15pt}
        \captionof{figure}{\textbf{Illustration of the \model}, physically interactable scene synthesis method to generate interactive 3D scenes characterized by \textbf{realistic layouts, articulated objects, and rich physical interactivity} tailored for embodied agents. }
\label{fig:overview}
\end{center}%
}]

\maketitle

\blfootnote{$^*$indicates equal contribution.}

\input{sec/0_abstract}    
\input{sec/1_intro}
\input{sec/2_related_work}

\input{sec/4_method}

\input{sec/5_experiment}
\input{sec/6_conclusion}

\clearpage

{
    \small
    \bibliographystyle{ieeenat_fullname}
    \bibliography{main}
}
\input{sec/8_appendix}


\end{document}

%% file: tex/preamble.tex
\usepackage[dvipsnames]{xcolor}
\newcommand{\red}[1]{{\color{red}#1}}
\newcommand{\green}[1]{{\color{ForestGreen}#1}}


%% file: tex/math_commands.tex
\usepackage{amsmath,amsfonts,bm}

\def\eqref#1{equation~\ref{#1}}

\def\1{\bm{1}}

\def\vb{{\bm{b}}}
\def\vc{{\bm{c}}}
\def\vd{{\bm{d}}}

\def\vo{{\bm{o}}}
\def\vp{{\bm{p}}}

\def\vx{{\bm{x}}}

\DeclareMathAlphabet{\mathsfit}{\encodingdefault}{\sfdefault}{m}{sl}
\SetMathAlphabet{\mathsfit}{bold}{\encodingdefault}{\sfdefault}{bx}{n}

\newcommand{\bx}{\mathbf{x}}

\newcommand{\bI}{\mathbf{I}}
\newcommand{\bg}{\mathbf{g}}
\newcommand{\btau}{\boldsymbol\tau}
\newcommand{\bmu}{\boldsymbol\mu}
\newcommand{\bepsilon}{\boldsymbol\epsilon}
\newcommand{\bSigma}{\boldsymbol\Sigma}

\newcommand{\cN}{\mathcal{N}}

%% file: tex/config.tex
\usepackage{acronym}

\definecolor{cvprblue}{rgb}{0.21,0.49,0.74}
\definecolor{purple}{rgb}{0.47, 0.32, 0.66}
\definecolor{blue}{rgb}{0.25, 0.41, 0.88}

\usepackage[pagebackref,colorlinks,breaklinks,citecolor=cvprblue]{hyperref}
\usepackage[misc]{ifsym}

\newcommand{\model}{\textsc{PhyScene}\xspace}
\newcommand{\sota}{state-of-the-art\xspace}

\acrodef{eai}[EAI]{embodied artificial intelligence}

\newcommand{\supp}{\textit{supplementary}\xspace}

\newcommand\blfootnote[1]{%
  \begingroup
  \renewcommand\thefootnote{}\footnote{#1}%
  \addtocounter{footnote}{-1}%
  \endgroup
}

%% file: sec/0_abstract.tex
\begin{abstract}
With recent developments in Embodied Artificial Intelligence (EAI) research, there has been a growing demand for high-quality, large-scale interactive scene generation. While prior methods in scene synthesis have prioritized the naturalness and realism of the generated scenes, the physical plausibility and interactivity of scenes have been largely left unexplored. To address this disparity, we introduce \model, a novel method dedicated to generating interactive 3D scenes characterized by realistic layouts, articulated objects, and rich physical interactivity tailored for embodied agents. Based on a conditional diffusion model for capturing scene layouts, we devise novel physics- and interactivity-based guidance mechanisms that integrate constraints from object collision, room layout, and object reachability. Through extensive experiments, we demonstrate that \model effectively leverages these guidance functions for physically interactable scene synthesis, outperforming existing \sota scene synthesis methods by a large margin. Our findings suggest that the scenes generated by \model hold considerable potential for facilitating diverse skill acquisition among agents within interactive environments, thereby catalyzing further advancements in embodied AI research.
\end{abstract}

%% file: sec/1_intro.tex
\section{Introduction}
\label{sec:intro}

The exploration of scene synthesis \cite{fu20213d,wang2021sceneformer,dhamo2021graph,zhai2023commonscenes,zhou2019scenegraphnet,chang2014learning,qi2018human,jiang2018configurable,fu2017adaptive,xu2013sketch2scene} has constituted a persistent focus within the field of computer vision. Initially conceived to facilitate indoor design applications, scene synthesis aimed to create diverse 3D environments characterized by both realism and naturalness. However, with the advent of \ac{eai}~\cite{ahn2022can,huang2022inner,driess2023palm,huang2023embodied}, the objectives of this task have taken on new dimensions. Simulated environments~\cite{kolve2017ai2,li2021igibson,szot2021habitat,xiang2020sapien,deitke2020robothor,deitke2022procthor}, now supporting a plethora of intricate embodied tasks, have propelled the task of scene synthesis into an important data source that provides unlimited scenarios for agents to robustly learn skills like navigation~\cite{anderson2018vision,krantz2020beyond} and manipulation~\cite{gong2023arnold, shridhar2022cliport, jiang2022vima}. This trend underscores the growing importance of scene synthesis within the context of \ac{eai} research.

Nevertheless, achieving a seamless transition from conventional scene synthesis algorithms to those tailored for \ac{eai} presents significant challenges in scene generation. As many \ac{eai} tasks involve physics simulation~\cite{mu2021maniskill,li2023behavior,mittal2023orbit,lin2021softgym,gu2023maniskill2,zheng2022vlmbench}, the synthesized scenes must adhere to physical constraints while enabling a high degree of interactivity among objects (\eg, articulated objects or fluids) and scene layout (\eg, reachability of objects) to facilitate agent skill acquisition. These stringent interactivity requirements introduce several obstacles for scene synthesis algorithms.
Limited by the quality of real-world scanned scenes~\cite{dai2017scannet,baruch2021arkitscenes,jia2024sceneverse}, previous methods have primarily relied on manually created scenes~\cite{fu20213dfuture,fu20213d}. However, these datasets are designed with non-interactable objects, overlooking physical constraints, and are prone to violations of such constraints. Consequently, this poses a significant challenge for algorithms aiming to learn physically plausible arrangements of interactable objects. Beyond data-level hurdles, incorporating scene interactivity (\eg, maintaining sufficient workspace, ensuring object reachability and interactivity) introduces non-trivial challenges in designing optimizable objectives that reflect such abstract concepts. These challenges emphasize the need for an effective scene synthesis algorithm that integrates the naturalness and realism of conventional synthesis algorithms while ensuring the physical plausibility and interactivity of scenes.


To address these challenges, we propose \model, \textit{a diffusion-based method embedded with physical commonsense for interactable scene synthesis.} Specifically, our approach builds on the efficacy of guided diffusion models~\cite{ho2022classifier,tang2023diffuscene,bansal2023universal,lu2022dpm} to effectively learn scene distribution and guide the model in generating scenes that are both functionally interactive and physically plausible. To incorporate articulated objects into generated scenes, we utilize the shape and geometry features, bridging rigid-body objects from training scenes with existing articulated object datasets. To model physical plausibility and interactivity accurately, we impose three key constraints on the generated scenes: (1) \textbf{physical collision avoidance} between objects to enable simulation, (2) \textbf{object layouts} constrained on the floor plan to avoid inter-room conflicts, and (3) the \textbf{interactiveness and reachability} of each object when assuming an embodied agent of proper size need to navigate. We convert these constraints into guidance functions that can be easily integrated into the guided diffusion model. We further propose metrics considering the aforementioned constraints in our evaluation process for assessing all existing models. Through meticulously designed experiments, we demonstrate that \model not only achieves \sota results on traditional scene synthesis metrics but also significantly enhances the physical plausibility and interactivity of generated scenes compared to existing methods. We hope this work can make a step forward in scalable indoor scene synthesis for \ac{eai} tasks, contributing to the broader landscape of \ac{eai} research. 

In summary, our main contributions are:
\begin{itemize}[leftmargin=*,nolistsep,noitemsep]
\item We propose \model, a guided diffusion model, for physically interactable scene synthesis with realistic layouts and interactable objects. 
\item Through well-crafted designs of guidance functions, we convert constraints encompassing collision avoidance, room layout, and reachability into \model in a simple and effective way to ensure the physical plausibility and interactivity of the generated scenes.
\item By comparing with competitive baseline models, we show that \model can not only achieve \sota results on traditional scene-synthesis metrics but also significantly outperforms existing methods for interactable scene synthesis on our delicately designed physical metrics, paving the way for new research topics bridging scene synthesis and \ac{eai}.
\end{itemize}

%% file: sec/2_related_work.tex
\section{Related Work}
\label{sec:related}

\paragraph{Indoor Scene Synthesis} 
Indoor scene synthesis is formulated as a layout prediction problem, where each object is often represented by its 3D bounding box, semantic labels~\cite{fu20213d,wang2021sceneformer}, or shape features~\cite{tang2023diffuscene} for retrieving corresponding meshes from 3D asset libraries to the specific locations.
To properly model the layout of objects in training datasets, current methods usually represent the arrangement of objects as a scene graph \cite{dhamo2021graph,zhai2023commonscenes,zhou2019scenegraphnet,chang2014learning} and utilize scene priors such as the spatial relationship between objects \cite{qi2018human} and object category (co-)occurrence frequency \cite{fu2017adaptive,xu2013sketch2scene} for approximating the scene layout distribution. While generating new scenes, these works leverage iterative sampling or optimization methods to reject scenes that violate the designed scene priors for synthesizing scenes with desired properties~\cite{fu2017adaptive,fisher2015activity,qi2018human,chang2014learning}. However, such methods are often limited by the efficacy of sampling or optimization algorithms. More recent works try to learn scene layout distributions with deep neural networks~\cite{paschalidou2021atiss,nie2023learning,wang2021sceneformer,huang2023diffusion,zhang2020deep,purkait2020sg,yang2021indoor} to improve the generation efficiency.

For the quality evaluation of generated scenes, common metrics test model performance with perceptual quality scores (\eg, FID~\cite{heusel2017gans}, KID~\cite{binkowski2018demystifying},\etc).
However, these realism metrics do not address the physical plausibility and interactivity of generated scenes, which is crucial for adapting scenes into simulated environments. In fact, a commonly used scene synthesis dataset, 3D-FRONT dataset~\cite{fu20213d}, exhibits frequent occurrence of these physically implausible layouts (as shown in \cref{tab:3dfront}). 
In addition, the interactivity of scenes for object manipulation and reachability is also understudied in prior works. 
ProcTHOR~\cite{deitke2022procthor} has proposed a procedural generation pipeline for interactable scenes with rule-based constraints and statistical scene priors. Nonetheless, as pointed out by~\cite{khanna2023habitat}, these generated scenes suffer from the pre-defined priors, thus generating unrealistic scenes that are harmful to agent learning. To this end, we aim to bridge this gap in \model, uniting efforts in scene synthesis and \ac{eai} to provide a pipeline that could suffice for large-scale interactable scene synthesis while maintaining visual realism and naturalness.

\paragraph{Physical Plausibility and Interactivity in 3D Scenes} Producing physically plausible generations in 3D scenes has been a long-standing problem for computer vision, given its subtleness in properly converting physical constraints into optimizable objectives. To tackle this challenge, various optimization-based approaches have been proposed for tasks such as scene-conditioned pose~\cite{hassan2019resolving} and motion generation~\cite{wang2021synthesizing}. However, the study of physical plausibility for scene generation has been largely left untouched. Meanwhile, the modeling of interactivity of 3D scenes has been largely left untouched in existing works without proper definition. 
Some works~\cite{wang2023rearrange} aim to define the level of scene interactivity via human and robot preferences in a scene rearrangement setting. Nonetheless, with their task-specific design, the optimization objectives are hard to be generalized to other settings. Therefore, \model aims at addressing these obstacles and makes the first attempts to provide reasonable definitions of physical plausibility and scene interactivity in the context of scene synthesis. 

\paragraph{Guided Diffusion Models}
Diffusion models~\cite{ho2020denoising,song2020denoising,lu2022dpm} have shown promising results for generative AI~\cite{ruan2023mm,inoue2023layoutdm,huang2023diffusion} across various domains~\cite{yu2022latent,poole2022dreamfusion,zhang2023adding,wu2023tune, wang2024move}. Through an iterative denoising process, diffusion models excel at handling high dimensional distributions without mode collapse. Such an iterative process also offers flexible ways to provide conditions~\cite{ramesh2022hierarchical,brooks2023instructpix2pix} and guidance \cite{ho2022classifier,bansal2023universal} that could effectively affect the inference of models. For example, SceneDiffuser \cite{huang2023diffusion} integrates a physics-based objective as conditional guidance for physically plausible planning and motion generation. PhysDiff \cite{yuan2023physdiff} proposes a physics-based motion projection module to instill the laws of physics into the denoising diffusion process for motion generation.
\model takes insight from these powerful techniques and integrates physical and interactivity guidance as conditional guidance for scene synthesis. 
Compared to constrained sampling methods such as Markov Chain Monte Carlo (MCMC) \cite{qi2018human,wang2023rearrange}, diffusion guidance runs more efficiently during the inference stage. Meanwhile, in contrast to models that take in constraints as a learnable objective~\cite{tang2023diffuscene}, our guidance functions can more effectively ensure the satisfaction of constraints during inference.
To the best of our knowledge, \model makes the first attempt to integrate a conditional diffusion model with physical plausibility and interactivity guidances to effectively generate physically interactable 3D scenes.

\begin{table}[t]
    \small
    \centering
    \caption{\textbf{Interactivity evaluation of scenes in the 3D-FRONT dataset.} These scenes exhibit a high rate of physical constraint violations including collision, layout, and interactivity. We provide detailed definitions of the metrics as explained in \cref{sec:exp}.}
    \vspace{-5pt}
    \resizebox{0.9\linewidth}{!}{
        \begin{tabular}{cccc}
            \toprule
            Data & Bedroom & Livingroom & Diningroom  \\
            \midrule
            $\text{Col}_\text{obj}\downarrow$ & 0.214   & 0.206  &  0.209  \\
            $\text{Col}_\text{scene}\downarrow$  & 0.42 &  0.625 &  0.57    \\
            $\text{R}_\text{out}\downarrow$   &  0.201 & 0.0584 & 0.159   \\
            $\text{R}_\text{walkable}\uparrow$   &  0.850 & 0.841 & 0.876   \\
            $\text{R}_\text{reach} \uparrow $  &  0.749 & 0.828 & 0.807   \\
            \bottomrule
        \end{tabular}
    }
    \label{tab:3dfront}
    \vspace{-12pt}
\end{table}

%% file: sec/4_method.tex
\section{\model}\label{sec:method}
\begin{figure*}[t!]
\centering
 \includegraphics[width=\linewidth]{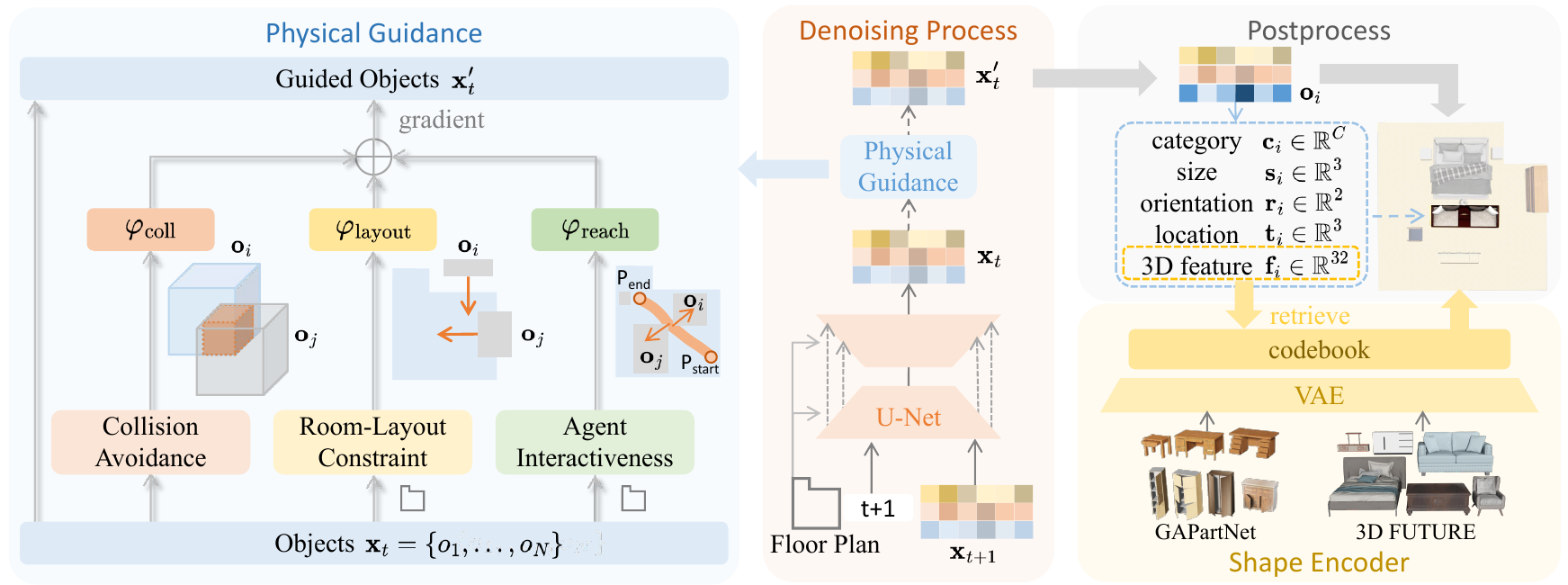}
\vspace{-15pt}
\caption{\textbf{Overview of \textsc{PHYSCENE}.} We leverage diffusion models for capturing scene layout distributions and apply three distinct guidance functions for improving the physical plausibility and interactivity of generated scenes.}
\label{fig:model_structure}
\vspace{-10pt}
\end{figure*}

Physically interactable scene synthesis requires realistic layouts, articulated objects, and physical interactivity. However, integrating articulated objects into scenes trained solely with static objects presents data-level challenges. We outline our method for incorporating articulated objects into generated scenes in~\cref{sub:feature_represent}. We then detail the model structure and training process of \model, where it learns prior layout knowledge from the dataset in~\cref{sub:training}.
To ensure physical interactivity, we consider collision avoidance, room layout 
constraint, and agent interactiveness as three key constraints, and provide details in~\cref{sub:guidance} on transforming them into guidance functions for posterior optimization during the inference process. 

\subsection{Object representation}
\label{sub:feature_represent}
The scene $\bx$ is composed of $N$ objects, noted as $\bx=\{\mathbf{o}_1,...,\mathbf{o_N}\}$. Each object representation $\mathbf{o}_i = [\mathbf{c}_i, \mathbf{s}_i, \mathbf{r}_i, \mathbf{t}_i, \mathbf{f}_i]$ is composed of a semantic label $\mathbf{c}_i\in \mathbb{R}^C$ out of C categories, size $\mathbf{s}_i\in \mathbb{R}^3$, orientation $\mathbf{r}_i = (cos\theta_i,sin\theta_i) \in \mathbb{R}^2$, location $\mathbf{t}_i\in \mathbb{R}^3$ and 3D feature $\mathbf{f}_i\in \mathbb{R}^{32}$ encoded from the shape of the object. Notably, common approaches for scene synthesis retrieve objects using the predicted size $\mathbf{s}_i$ and label $\mathbf{c}_i$. However, such methods could not be applied across asset libraries. We therefore leverage the shape feature $\mathbf{f}_i$ as a critical indicator for object retrieval, especially considering the objects in available articulated object datasets are largely different from those in scene synthesis datasets.
Specifically, we follow~\cite{tang2023diffuscene} and utilize a variational auto-encoder to embed object geometric features, transforming each 3D furniture model into a latent shape feature. For generating scenes with interactable objects, we consider object assets from: 1) 3D-FUTURE~\cite{fu20213dfuture} which contains CAD models used in 3D-FRONT~\cite{fu20213d}, and 2) GAPartNet~\cite{geng2023gapartnet} that includes various articulated objects. During inference, we use the latent encoded feature to find the best match of articulated objects in GAPartNet given the static objects in 3D-Front, thereby enabling the generation of scenes containing interactable objects.

\subsection{Conditional Diffusion for Layout Modeling}
\label{sub:training}
With a data sample $\bx_0$ representing the scene layout in the dataset, we gradually add Gaussian noise to $\bx_0$ with a forward process $q(\bx_{t+1}|\bx_t)$ converting it into a Gaussian noise $\bx_T$. Then a reverse denoising process $p_\theta(\bx_t|\bx_{t+1})$ is applied to recover the data from noise with learnable parameters $\theta$. Additionally, we consider using the floor plan $\mathcal{F}$ as a condition for incorporating the workspace and room layout constraints. In this case, we reconstruct $\bx_0$ via:
\begin{equation*}
    \begin{aligned}
        p_{\theta}(\bx_0 | \mathcal{F}) & = p(\bx_T)\prod_{t=1}^{T}p_{\theta}(\bx_{t-1}|\bx_{t}, \mathcal{F}), \\
        p_{\theta}(\bx_{t-1} | \bx_{t}, \mathcal{F}) & = \mathcal{N}(\bx_{t-1};\mu_\theta(\bx_{t}, t, \mathcal{F}), \Sigma_{\theta}(\bx_{t}, t, \mathcal{F})),
    \end{aligned}
\end{equation*}
where $p_{\theta}(\bx_0 | \mathcal{F})$ denotes the probability of scene layout $\bx_0$ given the conditional floor plan $\mathcal{F}$. As pointed out by previous works~\cite{ho2020denoising}, this maximization of conditional probability $p_\theta(\bx_0 | \mathcal{F})$ could be equivalently formulated as a simplified objective of estimating the noise $\epsilon$ through:
\begin{equation}
\small
\begin{aligned}
    \mathcal{L}_\theta(\bx_0|\mathcal{F}) & = \mathbb{E}_{t,\bepsilon,\bx_0}\left[\|\bepsilon - \bepsilon_{\theta}(\sqrt{\hat{\alpha}_t}\bx_0 + \sqrt{1 -\hat{\alpha}_t}\bepsilon,t,\mathcal{F})\|_2^2\right] \\
    & = \mathbb{E}_{t,\bepsilon,\bx_0}\left[\|\bepsilon - \bepsilon_{\theta}(\bx_t,t,\mathcal{F})\|_2^2\right],
\end{aligned}
\label{eq:conditional_ddpm}
\end{equation}
where $\hat{\alpha}_t$ is a pre-defined function of $t$ in the forward process according to a noise schedule (see details in the \supp).  To learn this conditional model, we utilize a U-Net with attention blocks to model $\bepsilon_\theta(\bx_t, t,\mathcal{F})$ with time embedding $t$ and floor plan embedding $\mathcal{F}$ added as conditions within every U-Net layer. 


\subsection{Guidance for Physical Interactivity}
\label{sub:guidance}
Considering the physical constraints violations in scenes from existing training data (as shown in \cref{tab:3dfront}), we ensure the physical plausibility and interactivity of generated scenes by guiding the conditional scene diffusion process with physic-based guidance functions. We start by first introducing guided sampling for diffusion models. Given constraint function $\varphi(\bx, \mathcal{F})$, we formulate the guided inference problem as optimizing the probability of constraint satisfaction:
\begin{equation}
    \begin{aligned}
    p(\bx_0| \mathcal{F}, O=1) & \propto p_{\theta}(\bx_0 | \mathcal{F})p(O=1|\bx_0, \mathcal{F})\\
    & \propto p_{\theta}(\bx_0 | \mathcal{F}) \cdot \exp\left(\varphi(\bx_0, \mathcal{F})\right),
    \end{aligned}
    \label{eq:opt_obj}
\end{equation}
where $O$ is an optimality indicator checking if the conditional generated output $\bx_t$ at denoising step $t$ satisfies the constraints in $\varphi(\bx, \mathcal{F})$. Similar to~\cite{huang2023diffusion}, we use the first order Taylor expansion around $\bx_t =\bmu$ at timestep $t$ to estimate the optimal condition in~\cref{eq:opt_obj} with:
\begin{equation}
    \begin{aligned}
    \log p(&O=1| \bx_t, \mathcal{F}) \approx (\bx_t - \bmu)\bg + C \\
    \bg & = \nabla_{\bx_{t}}\log p(O=1 | \bx_t, \mathcal{F})|_{\bx_t=\bmu} \\
    & = \nabla_{\bx_{t}}\varphi(\bx_t, \mathcal{F})|_{\bx_t=\bmu},
    \end{aligned}
\end{equation}
where $\bmu=\bmu_{\theta}(\bx_t,t,\mathcal{F})$, $\bg$ is the first order gradient estimate at $\bx_t =\bmu$ of $\log p(O=1|\bx_t, \mathcal{F})$, and $C$ is a constant. Therefore to generate a scene with constraints considered, we can modify the denoising process with a constraint perturbed Gaussian transition:
\begin{equation}
\begin{aligned}
    p_\theta(\bx_{t-1} | \bx_t, \mathcal{F}, O=1) = \mathcal{N}(\bx_{t-1}; \bmu + \lambda\bSigma\bg, \bSigma),
\end{aligned}
\label{eq:tilting}
\end{equation}
where $\bSigma = \bSigma_{\theta}(\bx_{t},t, \mathcal{F})$ and $\lambda$ is a scaling factor. Notably, the formulations in~\cref{eq:opt_obj} and~\cref{eq:tilting} leverage the predefined constraint functions $\varphi(\bx, \mathcal{F})$ as a tilting function on the original scene layout distribution to handle constraints. 

Under this formulation, we can easily combine the constraint functions into both learning and inference. Following~\cref{eq:conditional_ddpm}, we can reformulate the optimization of objective with $\varphi(\bx, \mathcal{F})$ through:
\begin{equation}
\small
    \begin{aligned}
        \mathcal{L}_{\theta}(\bx_0 | \mathcal{F}, O=1) = \mathbb{E}_{t, \bepsilon,\bx_0}\left[\|\bepsilon - \bepsilon_\theta(\bx_t, t,\mathcal{F}) - \lambda\bSigma\bg \|_2^2\right]
    \end{aligned}
\end{equation}
In scene synthesis, the guidance functions $\varphi(\bx_t, \mathcal{F})$ usually require real scene layouts for computing the violation constraints. Therefore, instead of optimizing for $\bx_t$ which might not be meaningful for real scenes, we convert the guidance functions into $\varphi(\tilde{\bx}_0^t,\mathcal{F})$ where $\tilde{\bx}_0^t$ is the predicted scene layout given initialization $\bx_t$. We summarize the guided learning and inference process of \model in~\cref{alg:diffuser:sample}.

\begin{algorithm}[t!]
    \small
    \caption{\small Learnning and inference in \model}
    \label{alg:diffuser:sample}
    \SetKwInOut{module}{Modules}
    \SetKwProg{Fn}{function}{:}{}
    \SetKwFunction{sample}{{\bf sample}}
    \module{Model $p_{\theta}(\cdot | \mathcal{F})$, guidance functions $\varphi(\cdot, \mathcal{F})=\{\varphi_{\text{coll}}(\cdot), \varphi_{\text{layout}}(\cdot, \mathcal{F}), \varphi_{\text{reach}}(\cdot, \mathcal{F})\}$.}
    \texttt{// constraint-guided learning}\\
    \KwIn{3D scene layout $\bx=\{\mathbf{o}_1,...,\mathbf{o}_N\}$ with floor plan $\mathcal{F}$, where N is a fixed number of objects.}
    \Repeat{\text{converged}}{  
        $\bx_0 \sim p(\bx_0|\mathcal{F})$ \\
        $\bepsilon \sim \mathcal{N}({\bf 0}, {\bf I})$, $t \sim \mathcal{U}(\{1,\cdots,T\})$\\
        $\bx_t = \sqrt{\hat{\alpha}_t}\bx_0 + \sqrt{1 - \hat{\alpha}_t}\bepsilon$, $\tilde{\bx}_0^t \sim p_{\theta}(\cdot|\mathcal{F})$\\
        $\theta = \theta - \eta\nabla_\theta \|{\bf\bepsilon} - {\bf\bepsilon}_{\theta}(\bx_t,t,\mathcal{F}) -\lambda\bSigma\bg\|^2_2$
    }
    \texttt{// one-step guided sampling}\\
    \Fn{\sample$(\btau^{t}$, $\varphi$)}{
        $\bmu = \bmu_{\theta}(\bx_t,t,\mathcal{F})$, $\bSigma = \bSigma_{\theta}(\bx_t,t,\mathcal{F})$\\
        $\varphi(\bx_t, \mathcal{F}) = \gamma_1\varphi_{\text{coll}}(\bx_t)+\gamma_2\varphi_{\text{layout}}(\bx_t,\mathcal{F})+\gamma_3\varphi_{\text{reach}}(\bx_t, \mathcal{F})$
        $\bx_{t-1} = \cN(\bx_{t-1}; \bmu + \lambda\bSigma\nabla_{\bx_t}\varphi(\bx_t,\mathcal{F})|_{\bx_t=\bmu}, \bSigma)$\\
        \Return $\bx_{t-1}$ \\
    }
    \texttt{// constraint-guided generation}\\
    \KwIn{initial scene layout $\bx_{T}\sim\cN({\bf 0}, {\bf I})$}
    \For{$t=T,\cdots,1$}{
        \texttt{// sampling with optimization}\\
        $\bx_{t-1}=\sample(\bx_{t}, \varphi)$\\
    }
    \Return $\bx_0$ \\
\end{algorithm}


 

Based on this formulation, we further propose three physic-based guidances $\varphi_{\text{coll}}(\bx)$, $\varphi_{\text{layout}}(\bx, \mathcal{F})$, and $\varphi_{\text{reach}}(\bx, \mathcal{F})$ and integrate them into the inference process as illustrated in~\cref{alg:diffuser:sample}. We detail the design of each guidance function as follows:

\textbf{Collision Avoidance}.
We design a collision avoidance function to reduce object mesh collisions in the generated scene. Instead of calculating the collision mesh between objects, we use the predicted bounding boxes and object centers as effective approximates for estimating the collision score of objects. Specifically, we use $\vb_i=[\mathbf{t}_i,\mathbf{r}_i,\mathbf{s}_i]$ to denote the 3D bounding box of object $\vo_i$ including its location $\mathbf{t}_i$, orientation $\mathbf{r}_i$ and size $\mathbf{s}_i$.
We use 3D IoU \cite{zhou2019iou} to calculate the collision guidance objective via:
\begin{equation}
    \varphi_{\text{coll}}(\bx) = -\sum_{i,j, i\neq j} \textbf{IoU}_{3D}(\vb_i, \vb_j),
\label{eq:coll_guide}
\end{equation}
where $\textbf{IoU}_{3D}$ represents the 3D bounding box IoU between object bounding boxes. We sum the collisions of each pair of objects in scene $\vx$ and take the negative value of the summation to penalize object collision.

\textbf{Room-layout guidance}
An important goal of scalable scene synthesis is to generate interactable house-level scenes in which embodied agents can navigate and interact. To achieve this goal, we consider adding the room-layout guidance that penalizes the existence of objects which are outside of a pre-given floor plan.
To consolidate this guidance function, we first extract a polygon of the room boundary given the floor plan $\mathcal{F}$. We then derive a set of $W$ outside barriers for identifying the boundary, represented as bounding boxes of walls $\{\mathbf{b}^{\text{wall}}_w\}_{w=1}^{W}$ with infinite thickness. We use a similar IoU score between objects and walls for room-layout guidance following:
\begin{equation}
    \varphi_{\text{layout}}(\bx|\mathcal{F}) =  -\sum_{i=1}^{N}\sum_{j=1}^{W} \mathbf{IoU}_{3D}(\vb_i,\vb_j^{\text{wall}}).
\end{equation}

\textbf{Reachability guidance} 
\label{sub:reachability_guidance}
For an embodied agent, the synthesized scene should allow it to traverse the entire room and interact with all objects successfully. Notably, the synthesized room is often separated into several disjoint connected regions in scenes with improper layouts. Based on this key observation, we aim to adjust the object locations that most significantly affect this connectivity between regions. More specifically, considering an embodied agent represented by its bounding box $\vb^{\text{agent}}$, we first map the generated scenes to a 2D room mask and calculate the walkable area in this scene considering the agent's size. Next, we employ Gaussian distributions on each positioned object in the scene to form a cost map for traversing the scene. Intuitively, points closer to objects will have higher costs. With the cost map, we plan the shortest path between the center of the two largest connection regions using the A* algorithm~\cite{Hart1968}. The resulting path indicates the least effort path to traverse between these two regions. We then select $L$ agent positions on this shortest path with bounding boxes $\{\vb^{\text{agent}}_1,...,\vb^{\text{agent}}_L\}$ for applying the guidance function. The reachability guidance can therefore be calculated via:

\begin{equation}
    \varphi_{\text{reach}}(\bx|\mathcal{F}) =  -\sum_{i=1}^{N}\sum_{j=1}^{L} \mathbf{IoU}_{3D}(\vb_i,\vb^{\text{agent}}_j).
\end{equation}

Notably, we can extend the current method to incorporate interaction constraints to ensure the articulated object interaction (\eg, grasping, opening) as well as complex rigid object interaction  (\eg, sit) with some simple modifications. More details are provided in the \supp.

%% file: sec/5_experiment.tex
\section{Experiment}\label{sec:exp}

\begin{table*}
\caption{\textbf{Quantitative comparison on unconditional scene synthesis trained on 3D-Front.} We compare \model with ATISS and Diffuscene on common perceptual quality scores FID, SCA, CKL, as well as physical plausibility measured in collision rate $\mathbf{Col}$.}
\vspace{-5pt}
\resizebox{\linewidth}{!}{
\begin{tabular}{cccccccccccccccc}
\toprule
\multirow{2}{*}{Method} & \multicolumn{5}{c}{Bedroom} & \multicolumn{5}{c}{Living Room} & \multicolumn{5}{c}{Dining Room} \\ \cmidrule(r){2-6} \cmidrule(r){7-11} \cmidrule(r){12-16} 
& FID $\downarrow$  & SCA $\downarrow$ & CKL $\downarrow$ & $\mathbf{Col}_{\text{obj}}$ $\downarrow$ & $\mathbf{Col}_{\text{scene}}$ $\downarrow$ & FID $\downarrow$  & SCA $\downarrow$ & CKL $\downarrow$ & $\mathbf{Col}_{\text{obj}}$ $\downarrow$ & $\mathbf{Col}_{\text{scene}}$ $\downarrow$ & FID $\downarrow$ & SCA $\downarrow$ & CKL $\downarrow$ & $\mathbf{Col}_{\text{obj}}$ $\downarrow$  & $\mathbf{Col}_{\text{scene}}$ $\downarrow$  \\ 
\midrule
ATISS~\cite{paschalidou2021atiss}&36.92& \textbf{49.24} &0.0036&  0.255 & 0.50& 55.76& 53.33 & 0.0016 &   0.372 & 0.870 &41.89 & 58.20 & 0.0028 & 0.483 & 0.91 \\
DiffuScene~\cite{tang2023diffuscene} &28.63 & 51.33&  0.0031   & 0.238  & 0.42 & 54.36& \textbf{50.24} & \textbf{0.0010}&  0.183  & 0.570 &  \textbf{37.68} & \textbf{57.60}    &  0.0031 &  0.253 & 0.63 \\
\midrule
PhyScene (Ours) &  \textbf{28.56} & 55.71  & \textbf{0.0030}  & \textbf{0.187}& \textbf{0.35}  & \textbf{40.67}   &  56.20   &  0.0015  & \textbf{0.130}    & \textbf{0.477}  &   37.88   &    58.74  & \textbf{ 0.0022}    & \textbf{0.134}    &  \textbf{0.40}\\
\bottomrule
\end{tabular}
}
    \label{tab:Unconditioned Scene Synthesis}
\end{table*}

\begin{figure*}[t!]
\centering
\includegraphics[width=1\linewidth]{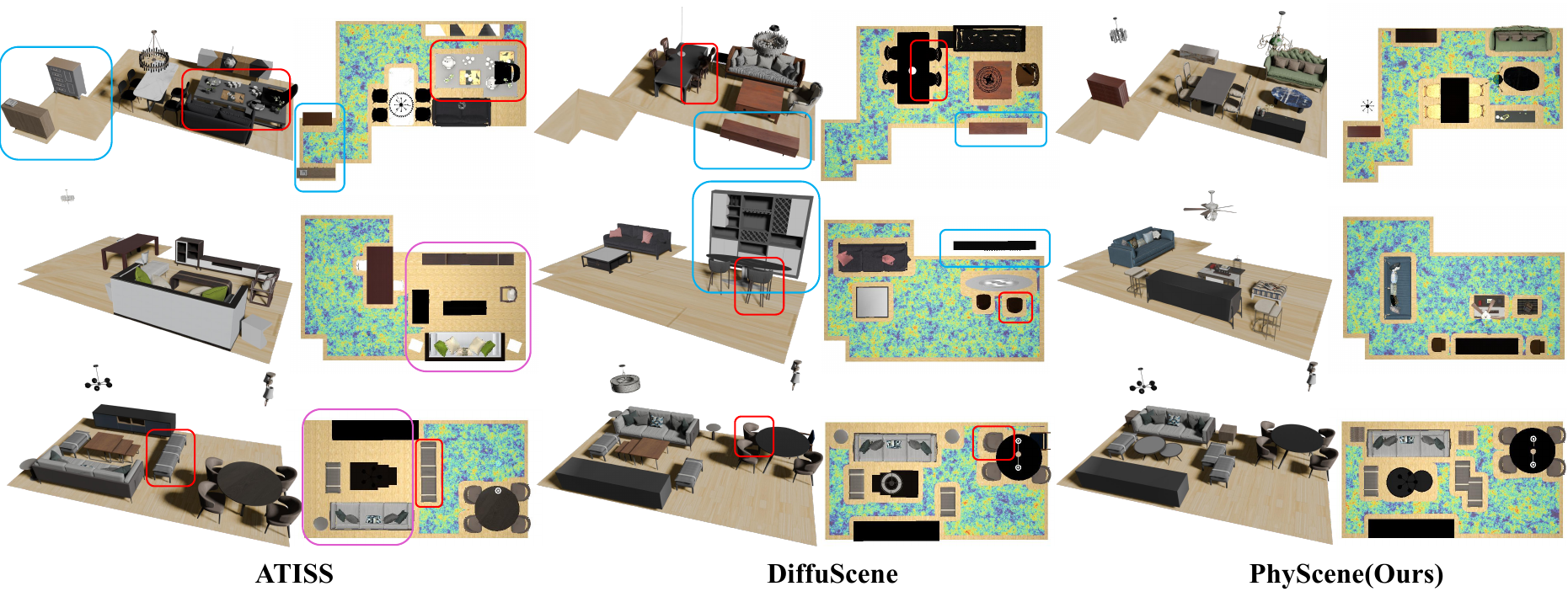}
\vspace{-15pt}
\caption{\textbf{Visualization of floor-plan conditioned scene synthesis between PhyScene, ATISS, and DiffuScene.} The \textbf{\red{red}}, \textbf{\textcolor{purple}{purple}}, and \textbf{\textcolor{blue}{blue}} boxes highlight collisions between objects, objects outside the floor plan, and unreachable areas to the embodied agent, respectively. }
\label{fig:Conditional_result}
\end{figure*}

\paragraph{Dataset}
For experimental comparisons, we train our diffusion model on the 3D-FRONT dataset \cite{fu20213d} which contains $6813$ houses with $14629$ rooms. Each room is manually decorated with high-quality furniture objects from the 3D-FUTURE dataset \cite{fu20213dfuture}. Following the setting of DiffuScene \cite{tang2023diffuscene} and ATISS \cite{paschalidou2021atiss}, we use 4041 bedrooms, 900 dining rooms, and 813 living rooms for training and testing. In addition, we use both the 3D-FUTURE dataset \cite{fu20213dfuture} and GAPartNet \cite{geng2023gapartnet} for object retrieval. Among them, GAPartNet \cite{geng2023gapartnet} has abundant interactive assets, containing $1166$ articulated objects from $27$ object categories. We utilize articulated objects in the table and storage furniture category, such as wardrobe and table, to retrieve related objects in generated scenes, and provide the full object category mapping between datasets in the \supp.
\paragraph{Baseline} We mainly consider two state-of-the-art scene synthesis methods as baselines: 1) ATISS \cite{paschalidou2021atiss}, a transformer-based model that predicts the 3D object bounding box in an autoregressive manner, and 2) DiffuScene \cite{tang2023diffuscene}, a diffusion-based model that learns 3D objects layout without floor plan constraint. We test these baselines in both unconditional synthesis and floor-plan-conditioned synthesis settings to compare our proposed \model model.

\paragraph{Metric} To evaluate the realism and diversity of the synthesized scenes, we follow the previous works and calculate Fr\'{e}chet Inception Distance \cite{heusel2017gans} (FID), Kernel Inception Distance \cite{binkowski2018demystifying} (KID $\times 0.001$), Scene Classification Accuracy (SCA), and Category KL divergence (CKL $\times 0.01$) on 1000 synthesized scenes.
In addition, we check the collision rate between each pair of objects in the generated scene using their CAD models. We use $\mathbf{Col}_{\text{obj}}$ to denote the percentage of objects that collide with other objects in the generated scene, $\mathbf{Col}_{\text{scene}}$ to denote the ratio of scenes that possess object collisions over all generated scenes.
Since the CAD models in the 3D-FUTURE dataset are usually not watertight, we apply re-meshing for each object mesh before evaluation.
To evaluate the violation of the floor plan layout, we mark the rate of objects outside the floor plan as $\mathbf{R}_{\text{out}}$. Finally, we calculate the average reachable rate of objects in the scene $\mathbf{R}_{\text{reach}}$ starting from a random starting point on the floor plan. We calculate the average ratio of the largest connected walkable area over all walkable areas in the room, denoted as $\mathbf{R}_{\text{walkable}}$, to evaluate the reachability and interactivity of the generated scenes.

\begin{table*}
\caption{\textbf{Floor-conditioned Scene Synthesis. } We compare \model with ATISS and DiffuScene on common perceptual quality scores FID, KID, SCA, CKL, as well as physical plausibility metrics $\mathbf{Col}_{\text{obj}}, \mathbf{Col}_{\text{scene}}, \mathbf{R}_{\text{out}}, \mathbf{R}_{\text{reach}}, \mathbf{R_{\text{walkable}}}$.}
\centering
\vspace{-5pt}
\resizebox{\linewidth}{!}{
    \begin{tabular}{ccccccccccc}
        \toprule
              Room Type & Method & FID $\downarrow$ & KID $\downarrow$ & SCA $\downarrow$ & CKL $\downarrow$ & $\mathbf{Col}_{\text{obj}}$ $\downarrow$ & $\mathbf{Col}_{\text{scene}} \downarrow$  & $\mathbf{R}_{\text{out}}$ $\downarrow$ & $\mathbf{R}_{\text{walkable}}$ $ \uparrow$ & 
            $\mathbf{R}_{\text{reach}}$ $\uparrow$\\ 
        \toprule
          
          \multirow{3}{*}{Bedroom} & ATISS  &  30.19  &  0.0010  & 49.14& 0.0028 & 0.248  &  0.46   & 0.286  &  0.839  &  0.736    \\
          & DiffuScene  & \textbf{25.00} & \textbf{0.0004}  & 51.78 &0.0031 & 0.228   & 0.43   & 0.272 &   0.827 &  0.755     \\
        \cmidrule(r){2-11}
          & PhyScene (Ours) & 25.52 & 0.0006 & \textbf{50.10} & \textbf{0.0025} & \textbf{0.187}  &  \textbf{0.36}  &  \textbf{0.245}     &  \textbf{0.865} &   \textbf{0.762}  \\
        \toprule
         \multirow{3}{*}{Living Room} & ATISS       &   45.66   &  0.0035  &  \textbf{51.64}   &  0.0016   &  0.316  &  0.85   & \textbf{ 0.136}  &   0.814   & \textbf{0.791 }     \\
           & DiffuScene  &   \textbf{38.69}   & \textbf{0.0012}   &  54.06  & 0.0017   &  0.198  &    0.69  &   0.238 &   0.790   &  0.756     \\
        \cmidrule(r){2-11}
           & PhyScene (Ours)    & 43.33  &  0.0031    &   53.50  & \textbf{0.0015}  & \textbf{0.191}   &  \textbf{0.63} & 0.219  & \textbf{0.815}     & 0.771     \\
        \toprule
           \multirow{3}{*}{Dining Room} & ATISS       &  41.66  &  0.0039  &  64.57& 0.0040& 0.591  & 0.96  & \textbf{0.132}    &   \textbf{0.874}   &  \textbf{0.848}   \\
          & DiffuScene  &  \textbf{38.31}  & \textbf{0.0020} &60.19 & \textbf{0.0013}   & 0.160 & 0.55 &  0.244  &    0.787 &  0.847     \\
        \cmidrule(r){2-11}
           & PhyScene (Ours) & 39.90 & 0.0026 & \textbf{60.00} & \textbf{0.0013} &  \textbf{0.151}   & \textbf{0.53}   & 0.217     &   0.852 &   0.789 \\
        \bottomrule
    \end{tabular}
}

\label{tab:conditioned Scene Synthesis}
\end{table*}

\subsection{Unconditioned Scene Synthesis} 
We provide quantitative evaluation results in~\cref{tab:Unconditioned Scene Synthesis}. As shown in \cref{tab:Unconditioned Scene Synthesis}, \model achieves \sota results on almost all metrics, especially with a significant improvement on physical plausibility metrics such as $\mathbf{Col}_{\text{obj}}$ and $\mathbf{Col}_{\text{scene}}$. This result quantitatively proves that \model effectively produces improved scene layouts with reduced collision rates while achieving better visual plausibility. Notably, diffusion-based scene-synthesis models (\ie, DiffuScene and \model) exhibit superior performance in collision avoidance compared to ATISS. This affirms the advantage of employing diffusion models as the primary generative model for scene synthesis, given their robust performance and adaptability in integrating guidance functions. We provide qualitative results in~\cref{fig:Conditional_result}, demonstrating that our model successfully generates scenes with significantly fewer instances of physical constraint violations due to object collisions while maintaining high levels of naturalness and diversity.

\begin{figure}[t!]
\centering
\includegraphics[width=\linewidth]{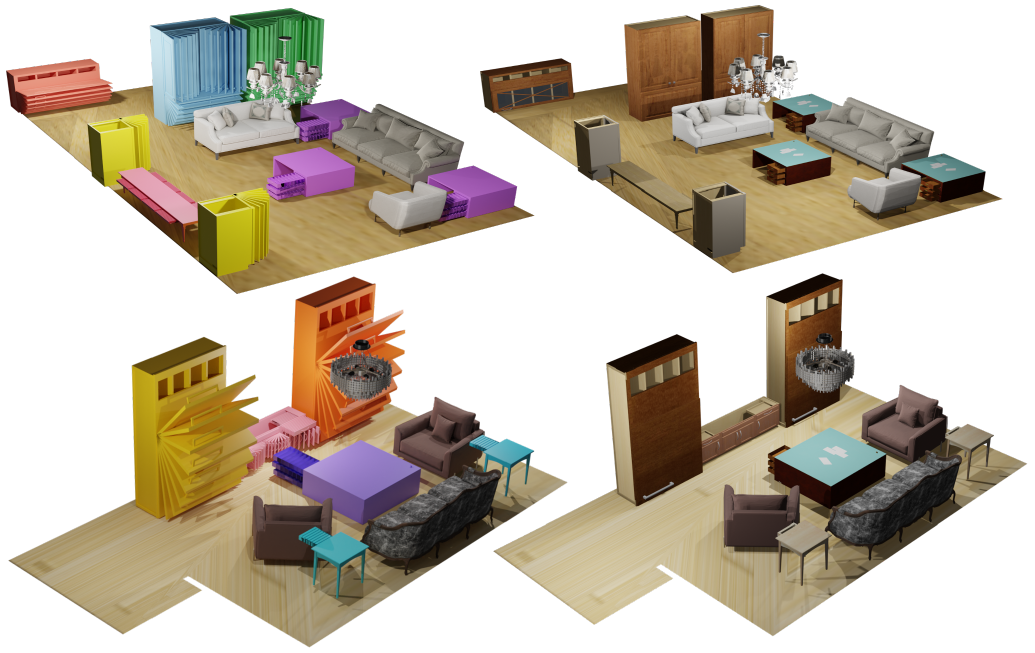}
\caption{\textbf{Generated scenes with articulated objects.} We visualize the opening sequence of articulated objects (left) and the generated scenes with texture (right).}
\label{fig:articulation}
\vspace{-10pt}
\end{figure}

\subsection{Floor-conditioned Scene Synthesis}
We provide comparisons between \model and baseline models in terms of both visual and physical metrics in \cref{tab:conditioned Scene Synthesis}. 
\model surpasses baselines in collision metrics and the CKL score. Additionally, compared to DiffuScene, our model consistently exhibits performance improvements across all physical interactability metrics, highlighting the effectiveness of our physical guidance functions in enhancing the generation process of diffusion-based models with physical constraints. It is noteworthy that, except for the Bedroom setting, ATISS achieves favorable results on floor plan violation ($\mathbf{R}_{\text{out}}$) and reachability metric ($\mathbf{R}_{\text{reach}}$). We attribute this to its prioritization of floor plan constraints over collision avoidance within the scene. We provide qualitative visualization of all models' generations in~\cref{fig:Conditional_result}.

\begin{table}
\centering
    \caption{\textbf{Articulated Object Embedding.} We compare \model with ATISS and DiffuScene on physical plausibility.}
   \vspace{-5pt}
   \resizebox{\linewidth}{!}{
        \begin{tabular}{ccccc}
            \toprule
            Method & $\mathbf{Col}_{\text{obj}}$ $\downarrow$ & $\mathbf{Col}_{\text{scene}} \downarrow$ & $\mathbf{R}_{\text{out}}$ $\downarrow$ & $\mathbf{R}_{\text{reach}}\uparrow$ \\
             \midrule
            ATISS & 0.360 &  0.86  & \textbf{0.154} & \textbf{0.758} \\
            DiffuScene & 0.262  & 0.78  & 0.237 & 0.702 \\
            PhyScene (Ours) & \textbf{0.251} & \textbf{0.76} & 0.229 & 0.755\\
            \bottomrule
        \end{tabular}
        }
    \label{tab:Collision rate of articulated Scene Synthesis}
\end{table}

\subsection{Scene Synthesis with Articulated Objects}
To generate scenes with articulated objects, we utilize the predicted scene layout along with object features to retrieve articulate objects. Recognizing the spatial requirements for interacting with articulated objects, we compute 3D bounding boxes for these objects, considering their joints being manipulated to the fullest extent, and use these expanded bounding boxes for guidance calculation. 
We show the quantitative results of our guided substitution for articulated objects in the Living Room setting in \cref{tab:Collision rate of articulated Scene Synthesis}.
Results show the collision rate with articulated objects is much higher than that with rigid objects (compared with the collision rate shown in \cref{tab:conditioned Scene Synthesis}). And our model shows a great improvement over previous methods.
We visualize the qualitative results of our guided substitution for articulated objects in ~\cref{fig:articulation} and leave more visualizations in the \supp.



\begin{table}[t!]
\caption{\textbf{Ablation study on the use of guidance functions.} Our final result balances the effectiveness of three guidances.}
\vspace{-5pt}
\resizebox{\linewidth}{!}{
\begin{tabular}{ccccccc}
\toprule
 Collision & Layout & Interact  &  $\mathbf{Col}_{\text{obj}}\downarrow$ & $\mathbf{R}_{\text{out}}\downarrow$        & $\mathbf{R_{\text{walkable}}}\uparrow$     & $\mathbf{R}_{\text{reach}}\uparrow$    \\ \midrule
            &            &            & 0.200 & 0.240 &0.808 &  0.763     \\
 \checkmark &            &            & 0.111 &   0.354 & 0.832 &  0.793 \\
            & \checkmark &            &  0.279 & 0.110 &0.774    &0.742   \\
            &            & \checkmark &  0.239  & 0.260  &0.927  & 0.813  \\ \midrule
 \checkmark & \checkmark & \checkmark & 0.191 &  0.219   & 0.815 & 0.771  \\
 \bottomrule
\end{tabular}
}
\label{tab:ablation study}
\end{table}

\subsection{Ablation Study on Guidance} 
\begin{figure*}[t!]
\centering
\includegraphics[width=1.0\linewidth]{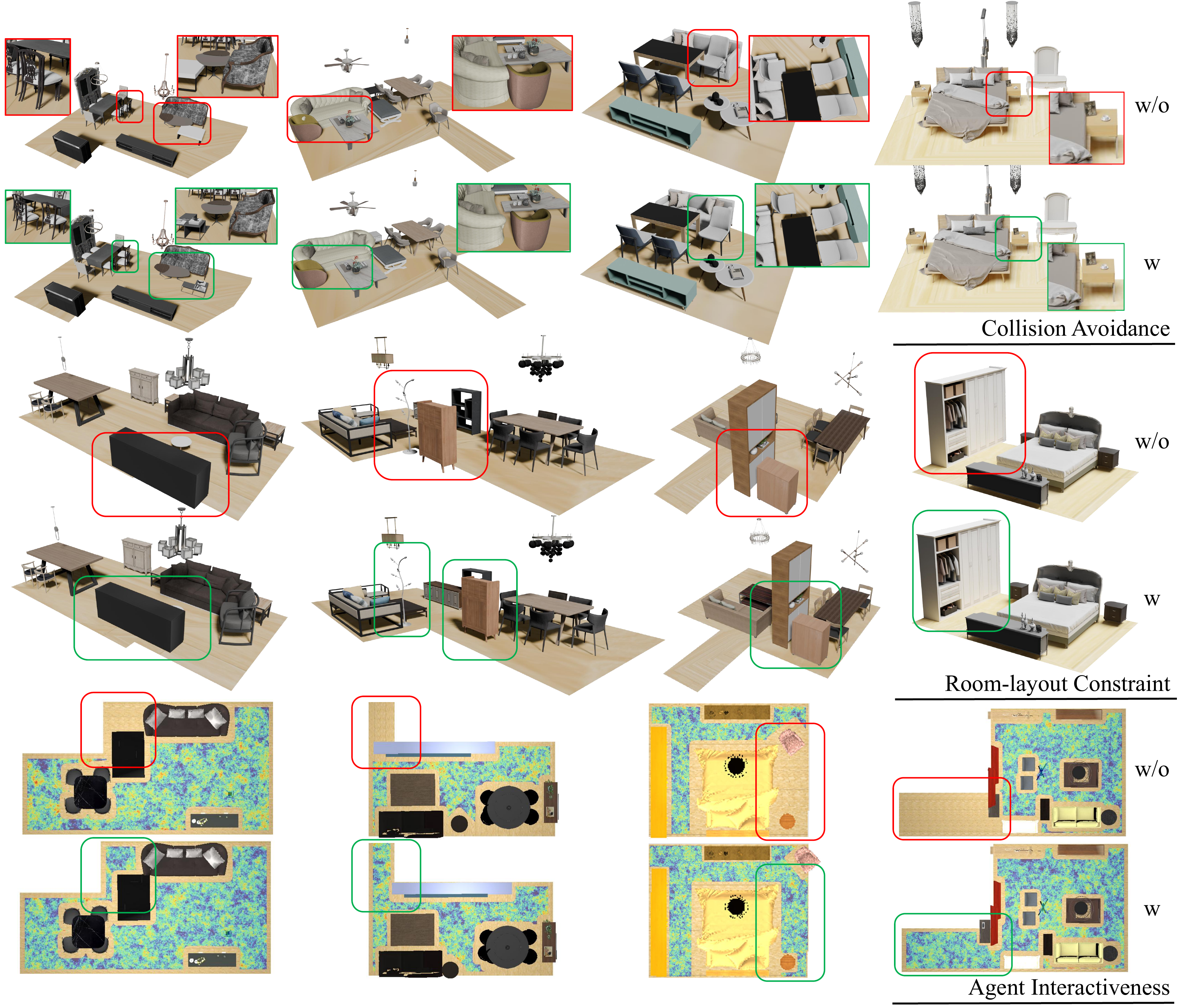}
\vspace{-15pt}
\caption{\textbf{Ablation on Guidance.} Results of different guidance with floor-plan conditions. For each ablation on guidance functions, we show four generated scenes (four columns) without guidance in the first row and mark the violation of constraints in \textbf{\red{red}} boxes. The second row shows the improvement after considering guidance functions in \textbf{\green{green}} boxes. }
\vspace{-10pt}
\label{fig:guidance}
\end{figure*}

We conduct ablative studies on our proposed guidance functions in the Living Room setting in~\cref{tab:ablation study}. Given that these guidance functions serve different roles in layout optimization, they may exhibit potential conflicts with each other. As shown in~\cref{tab:ablation study}, the $\mathbf{Col}_\text{obj}$ and $\mathbf{R}_\text{out}$ metrics have a negative impact on each other because the collision guidance pushes objects apart while the floor plan guidance pushes objects closer to fit in the scene. However, we managed to strike a balance among these guidances, leading to improvements on each corresponding metric. We provide qualitative visualizations illustrating the effect of each guidance in \cref{fig:guidance}.

%% file: sec/6_conclusion.tex
\section{Conclusion}

In this paper, we introduce \model, a guided conditional diffusion model for physically interactable scene synthesis. To ensure the physical plausibility and interactivity of the generated scenes, we devise novel guidance functions converting constraints on object collision, room layout, and interactivity to guidance within each inference step in the diffusion process. Our experimental results demonstrate consistent performance improvement over \sota baseline models on physical plausibility and interactivity metrics, showcasing the effectiveness of our designed guidance functions and the generation pipeline.
\vspace{5pt}

\noindent \textbf{Future work} Due to data limitations, \model is presently restricted to considering only limited room types, without incorporating small objects. This limitation poses a significant obstacle to the applicability of these scenes in embodied AI tasks, particularly those involving small object manipulation such as pick and place tasks. We leave this area as an important focus for future research.
\vspace{5pt}

\noindent \textbf{Acknowledgement} We
thank Ms. Zhen Chen from BIGAI for refining the figures, and all colleagues from the BIGAI TongVerse project for fruitful discussions and help on simulation developments. We would also like to thank the anonymous reviewers for their constructive feedback.

%% file: sec/8_appendix.tex
\clearpage
\setcounter{page}{1}
\renewcommand{\thefigure}{A.\arabic{figure}}
\renewcommand{\thetable}{A.\arabic{table}}
\renewcommand{\theequation}{A.\arabic{equation}}
\appendix
\setcounter{figure}{0}
\setcounter{table}{0}
\setcounter{equation}{0}
\maketitlesupplementary

\section{Algorithm Details}
\subsection{Details of Parameters }
We introduce the details of $\hat{\alpha}_t$ in \cref{eq:conditional_ddpm}.
Given a data sample $\bx_0$, we can define a forward diffusion process by adding noise. Each forward diffusion process adds Gaussian noise with variance $\beta_t$ on $\bx_{t-1}$, resulting in a new variable $\bx_t$ with distribution $q(\bx_t|\bx_{t-1})$. This process can be formulated as:

\begin{equation*}
\begin{aligned}
    q(\bx_t|\bx_{t-1}) = \mathcal{N}(\bx_t; \bmu_t=\sqrt{1-\beta_t}\bx_{t-1},\bSigma_t=\beta_t\bI).
\end{aligned}
\label{eq:sforward1}
\end{equation*}

Then we can formulate the diffusion process with

\begin{equation*}
\begin{aligned}
    q(\bx_{1:T}|\bx_{0}) = \prod_{t=1}^Tq(\bx_{t}|\bx_{t-1}),
\end{aligned}
\label{eq:sforward2}
\end{equation*}
where $q(\bx_{1:T})$ means we apply $q$ repeatedly from timestep 1 to $T$. To simplify this process, we define $\alpha_t=1-\beta_t$, $\hat{\alpha}_t=\prod_{s=0}^t\alpha_s$, and $\bepsilon, \bepsilon_{0}, ..., \bepsilon_{t-1} \sim \mathcal{N}(0,\bI)$.
After reparameterizing with $\hat{\alpha}_t$, we have:

\begin{equation*}
\begin{aligned}
    \bx_t = & \sqrt{1-\beta_t}\bx_{t-1}+\sqrt{\beta_t}\bepsilon_{t-1} \\
          = & \sqrt{\alpha_t}\bx_{t-1}+\sqrt{1-\alpha_t}\bepsilon_{t-1} \\
          = & \sqrt{\alpha_t}(\sqrt{\alpha_{t-1}}\bx_{t-2}+\sqrt{1-\alpha_{t-1}}\bepsilon_{t-2})+\sqrt{1-\alpha_t}\bepsilon_{t-1}  \\
          = & \sqrt{\alpha_t\alpha_{t-1}}\bx_{t-2}+\sqrt{1-\alpha_t\alpha_{t-1}}\bepsilon \\ 
          = & ... \\
          = & \sqrt{\alpha_t\alpha_{t-1}...\alpha_1}\bx_0+\sqrt{1-\alpha_t\alpha_{t-1}...\alpha_1}\bepsilon \\ 
          = & \sqrt{\hat{\alpha}_t}\bx_{0}+\sqrt{1-\hat{\alpha}_t}\bepsilon.
\end{aligned}
\label{eq:sforward3}
\end{equation*}
This reflects the derivation between $x_t$ and $x_0$ in~\cref{eq:conditional_ddpm}.


\subsection{Details of Reachability Guidance} 

As mentioned in~\cref{sec:method}, we provide the detailed algorithm for calculating the reachability guidance in \cref{alg:reachable guide}.

\begin{algorithm}[t!]
    \small
    \caption{\small Reachability Guidance}
    \label{Reachability Guidance}
    \SetKwInOut{module}{Module}
    \SetKwProg{Fn}{function}{:}{}
    \SetKwFunction{sample}{{\bf sample}}
    \module{Reachability guidance function $\varphi_{\text{reach}}(\cdot|\mathcal{F})$, search algorithm $\textbf{A}^*(\cdot)$, indicator function $\mathbbm{1}(\cdot)$.}
    \KwIn{Floor plan $\mathcal{F}$, 3D object bboxes $\{\vb_{1},...,\vb_{N}\}$ where N is the number of objects, embodied agent 's width $\vd$.}

    \texttt{//Generate gaussian cost map}\\
    $W = \mathbbm{1}(\mathcal{F})$  \texttt{//Init walkable area} \\
    $C = \neg\mathbbm{1}(\mathcal{F})\cdot \text{MAX\_VALUE}$  \texttt{//Init cost map}\\
    
    \For{$i=1,\cdots,N$}{
        $\vb^{2D}_{i} = \textbf{\textsc{MapTo2D}}(\vb_{i})$\\
        $W = W - \mathbbm{1}(\textbf{\textsc{Dilate}}(\vb^{2D}_{i}, \vd/2))$\\
        \texttt{//Add Gaussian cost for each object} \\
        $C = C + \textbf{\textsc{Gaussian}}(\vb^{2D}_{i})$  \\ 
    }
    
    \texttt{//$A^*$ shortest path search}\\
    $\{\vc_1, ...,\vc_M\} = \textbf{\textsc{FindConnectedArea}}(W)$\\
    $\{\vp_1, ...,\vp_M\} = \textbf{\textsc{FindCenter}}(\{\vc_1, ...,\vc_M\})$\\
    \texttt{//Randomly choose $\vp_{start}$ and $\vp_{end}$}\\ 
    $\mathbf{Path}_\text{shortest} = \textit{\textbf{A}}^*(C, \vp_{start}, \vp_{end})$\\
    $\{\vb^{\text{agent}}_j\}_{j=1}^L = \textbf{\textsc{GetAgentBox}}(\mathbf{Path}_\text{shortest})$\\
    
    \texttt{// Reachability Guidance}\\
    $\varphi_{\text{reach}}(\bx|\mathcal{F}) =  -\sum_{i=1}^{N}\sum_{j=1}^{L} \mathbf{IoU}_{3D}(\vb_i,\vb^{\text{agent}}_j)$ \\

    \Return $\varphi_{\text{reach}}(\bx|\mathcal{F})$ \\
    \label{alg:reachable guide}
\end{algorithm}

\begin{figure*}[t!]
\centering
\includegraphics[width=\linewidth]{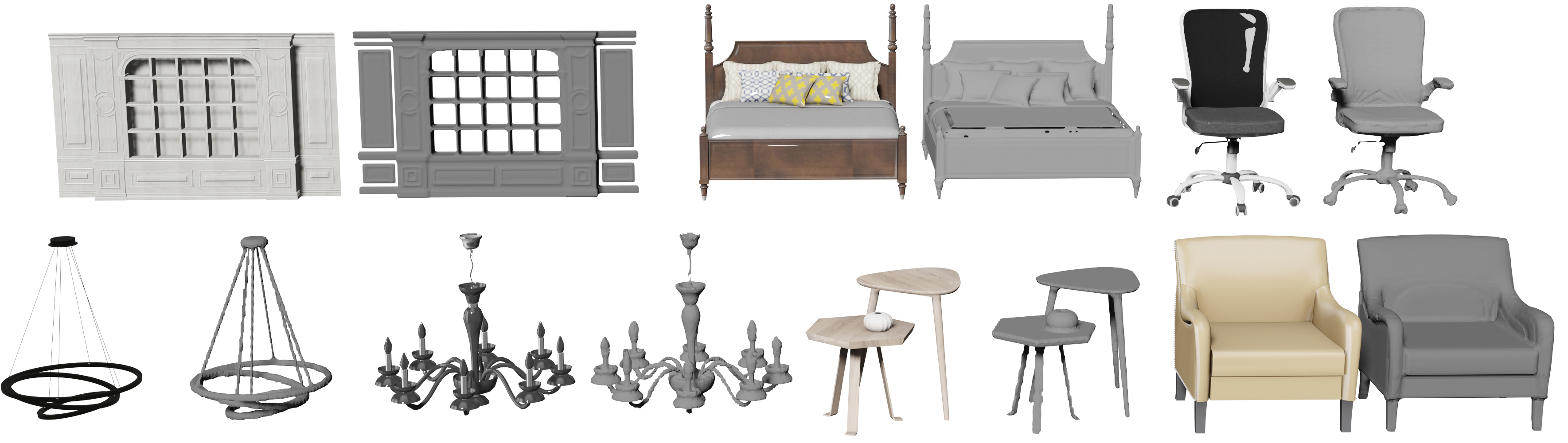}
\vspace{-10pt}
\caption{\textbf{Original 3D-FUTURE model v.s. re-meshed model.} We show examples of re-meshed models. Models on the left model are the original CAD model in 3D-FUTURE, and on the right are the re-meshed models. Despite the perceptual similarity, the re-meshed models fill in the hollow area for collision calculation.}
\label{fig:remesh}
\end{figure*}

\begin{figure*}[t!]
\centering
\includegraphics[width=\linewidth]{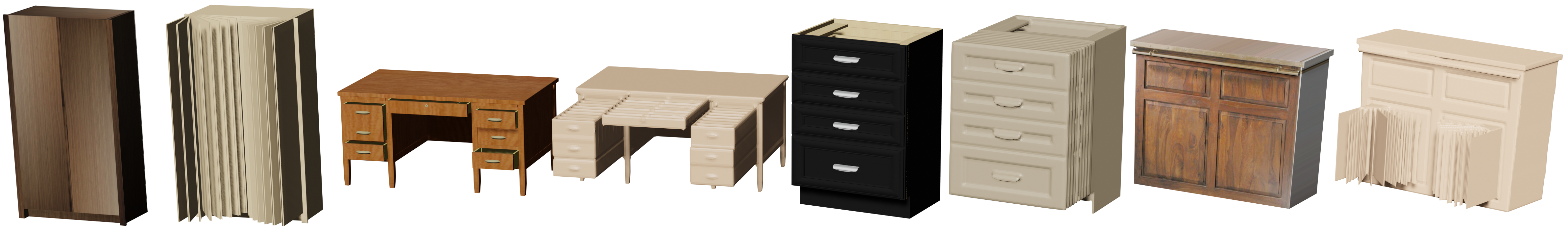}
\vspace{-10pt}
\caption{\textbf{Original GAPartNet model v.s. sequential model.} The original CAD models are always in closed status. To simulate the interactive situation, we open the furniture and record the sequential process in an integrated mesh. The left model shows the original furniture, while the right one is the sequential model. We use the sequential model to compute the collision rate of articulated objects. }
\label{fig:sequence}
\end{figure*}

\begin{figure*}[t!]
\centering
\includegraphics[width=\linewidth]{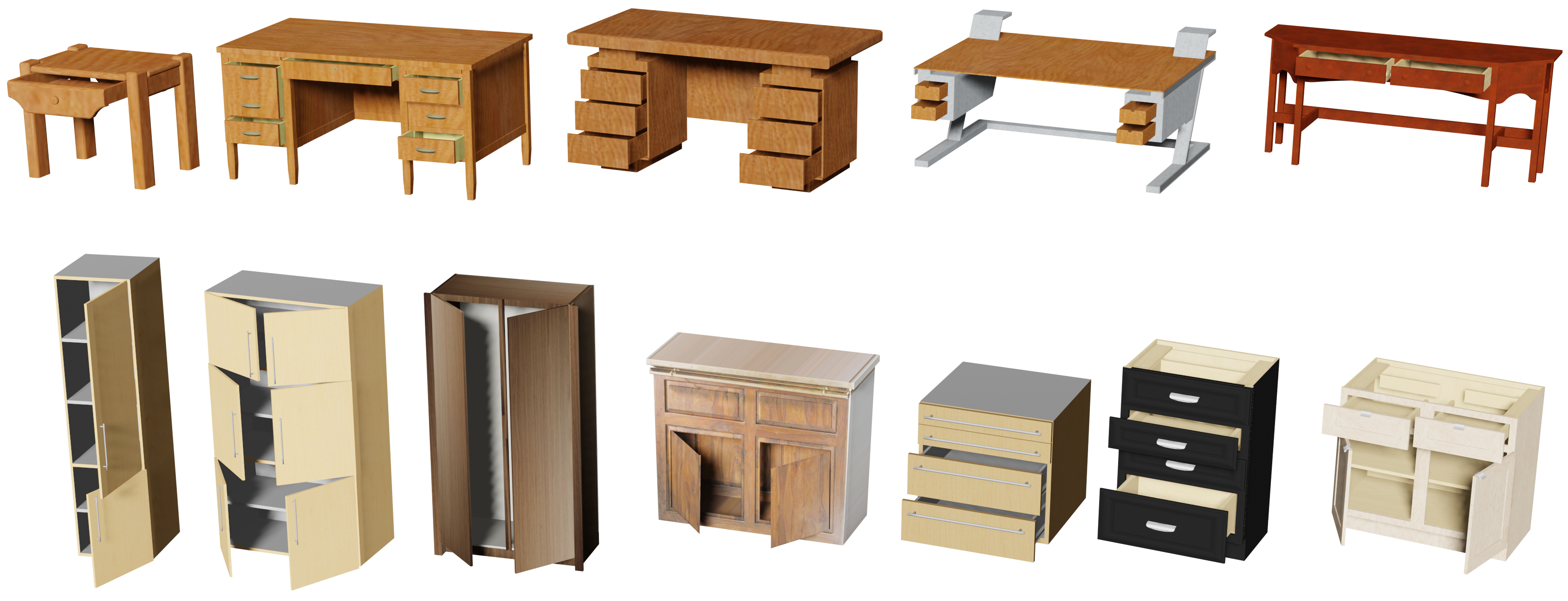}
\vspace{-10pt}
\caption{\textbf{Examples of articulated objects in GAPartNet dataset.} We visualize some models of \textit{StorageFurniture} and \textit{Table}. The articulated models have various appearances and different joint types such as revolute and prismatic. Each piece of furniture has several joints for interaction. }
\label{fig:ARTICULATED}
\end{figure*}








\section{Data Processing}
\subsection{3D-FUTURE} 
The original 3D-FUTURE dataset contains object CAD models that are not watertight, which can not be used for calculating collision directly. To solve this problem for evaluating physical collision between objects, we re-mesh each object model in Blender to compute the collision rate.
Some examples of re-meshed models are shown in \cref{fig:remesh}, where models on the left are original CAD models in 3D-FUTURE and those on the right are the re-meshed models. Despite the perceptual similarity between models provided and re-meshed, most provided samples contain hollows inside that forbid collision calculation.

\subsection{GAPartNet} To simulate the interaction between robots and articulated objects, we build upon the object CAD models and URDF files provided in GAPartNet. Specifically, we generate the articulated object's states from close to open according to the URDF file and record the sequential process into an integrated mesh. As shown in~\cref{fig:sequence}, we show the original object CAD model on the left and the integrated mesh covering articulated object states on the right. In our experiments, we use the integrated mesh to compute the collision rate between articulated objects and also use this integrated mesh to compute the opening size of articulated objects for guidance calculation.

\subsection{Retrieval Categories}
\begin{figure}[t!]
\centering
\includegraphics[width=1\linewidth]{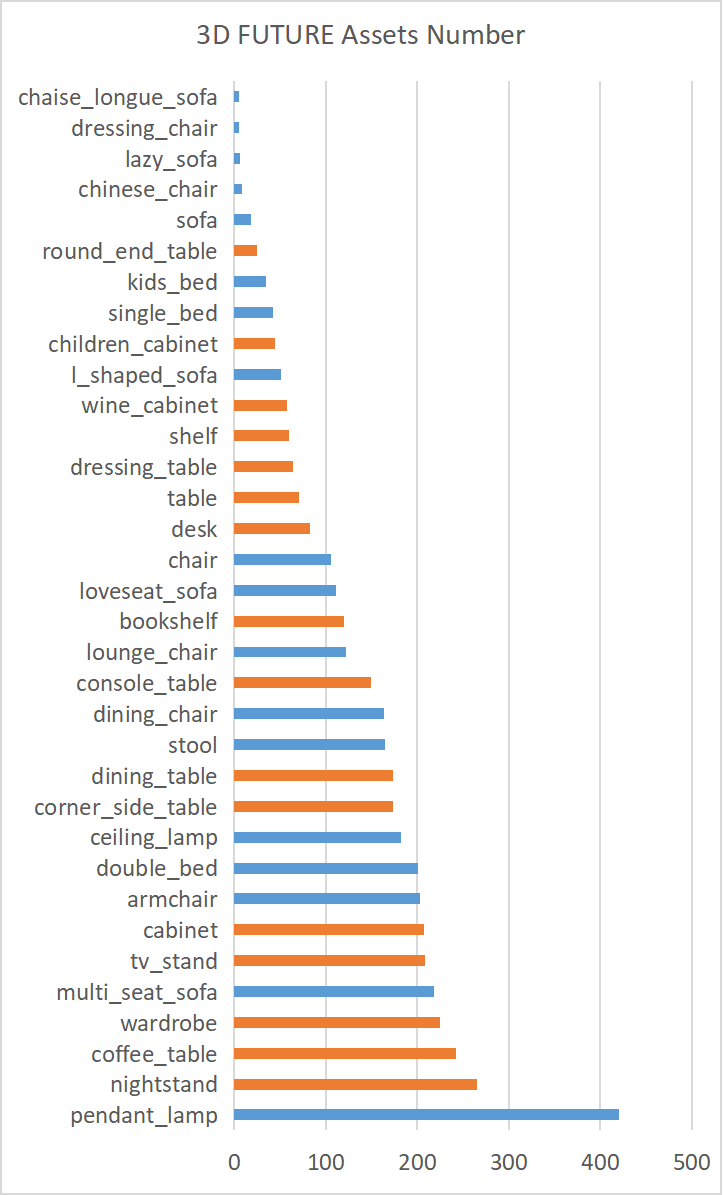}
\vspace{-10pt}
\caption{\textbf{Category distribution in 3D-FUTURE dataset.} We show the utilized categories in 3D-FUTURE dataset with asset numbers. We choose interactive categories such as \textit{wardrobe}, shown in orange, to retrieve GAPartNet model.}
\label{fig:3dfuture_category}
\end{figure}

\begin{figure}[t!]
\centering
\includegraphics[width=1\linewidth]{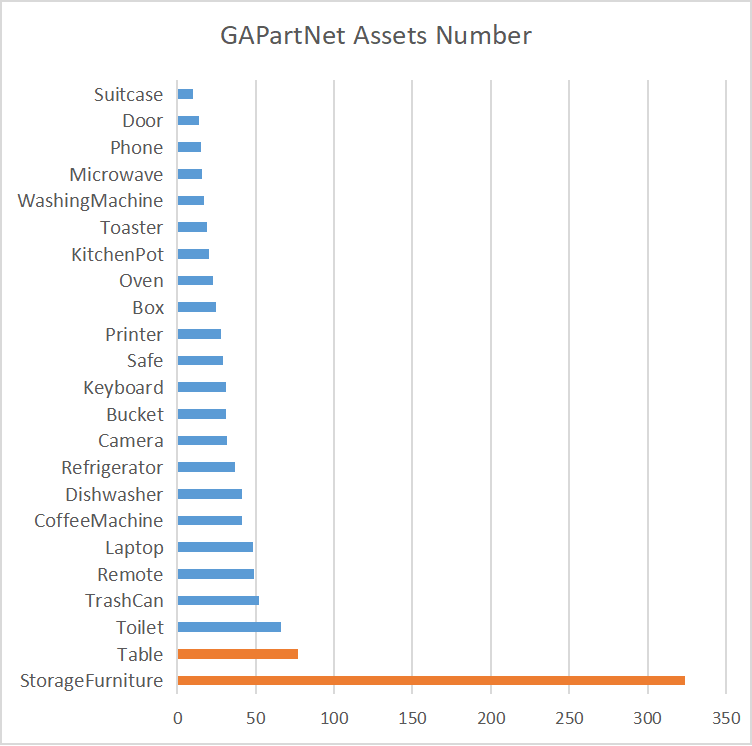}
\vspace{-10pt}
\caption{\textbf{Category distribution in GAPartNet dataset.} We show the category distribution of GAPartNet model, where \textit{StorageFurniture} and \textit{Table} take the largest proportion of this dataset. These two categories, as shown in orange, are used to composite interactable scenes with cross-dataset retrieval.}
\label{fig:gapartnet_category}
\end{figure}

As our method still primarily depends on retrieving object models for generating the final scene, we combine assets from the 3D-FUTURE and GAPartNet datasets for retrieval.
In \cref{fig:3dfuture_category} we show the utilized categories in the 3D-FUTURE dataset with their corresponding asset numbers. We build a mapping between the 3D-FUTURE object assets and GAPartNet to align interactive categories between two datasets, such as \textit{wardrobe} in the 3D-FUTURE, shown in orange, for the category of \textit{StorageFurniture} in the GAPartNet. \cref{fig:gapartnet_category} shows the category distribution of GAPartNet models, where \textit{StorageFurniture} and \textit{Table} take the largest proportion of this dataset. For example, the number of \textit{StorageFurniture} is 324 out of the whole dataset number 1045. The articulated models have various appearances and different joint types such as revolute and prismatic. Each piece of furniture has several joints for interaction.  We visualize some models of \textit{StorageFurniture} and \textit{Table} in GAPartNet in \cref{fig:ARTICULATED}.

\section{Additional Results}
\subsection{Physical Implausible Scenes in 3D-FRONT}
\begin{figure*}[t!]
\centering
\includegraphics[width=\linewidth]{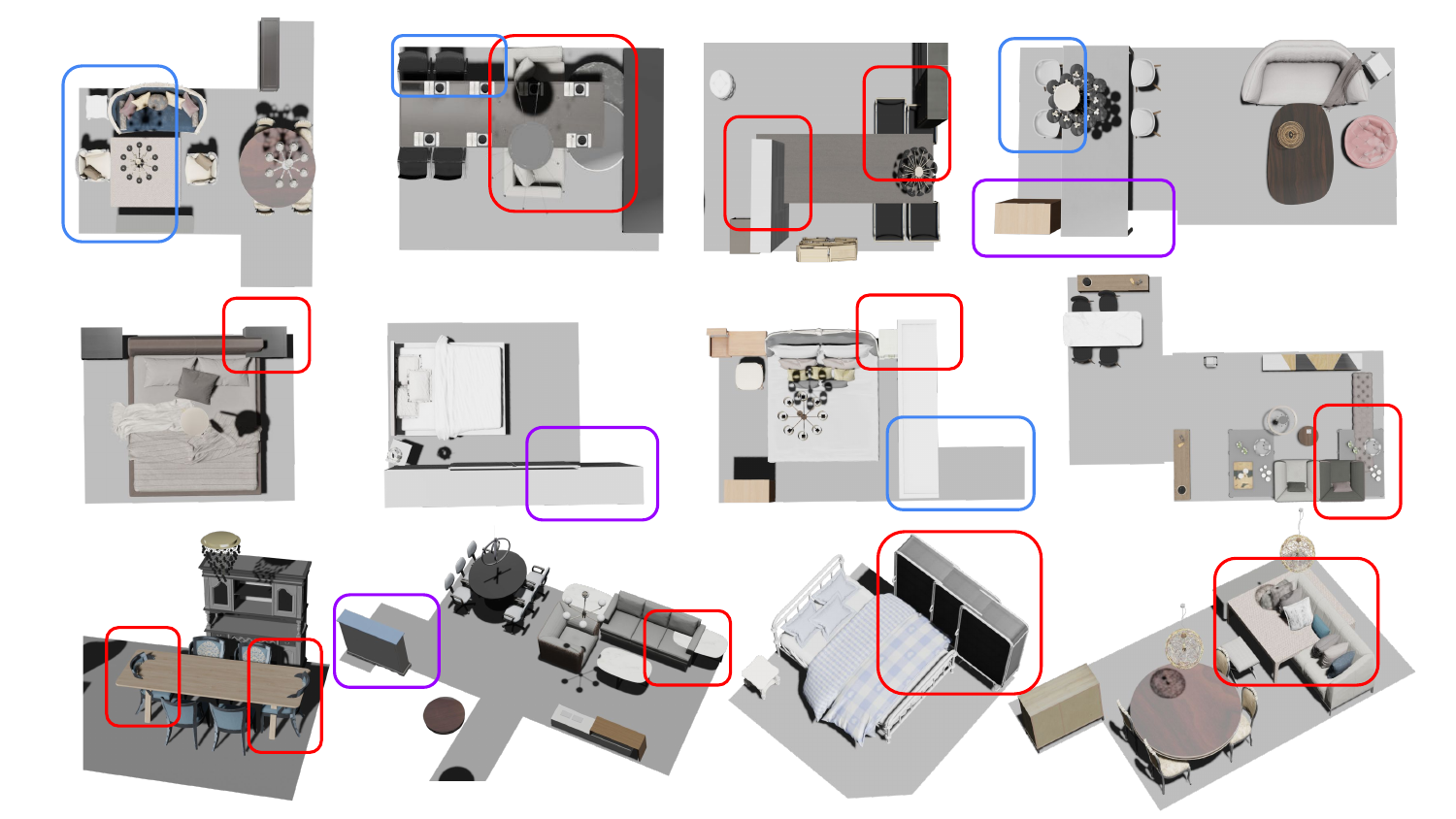}
\vspace{-15pt}
\caption{\textbf{Visualization on physically implausible scenes in 3D-FRONT.} We show original 3D-FRONT scenes with physical and interactive failure cases. The \textbf{\red{red}}, \textbf{\textcolor{purple}{purple}}, and \textbf{\textcolor{blue}{blue}} boxes respectively indicate collisions between objects, objects outside the floor plan and unreachable areas to the embodied agent. Here we set the floor plan in gray color without texture.}
\label{fig:3dfront}
\end{figure*}

As briefly discussed in~\cref{tab:3dfront}, we provide further qualitative visualizations on the violation of physical plausibility in 3D-FRONT scene data in~\cref{fig:3dfront}. As shown from the visualizations, some of the scenes used for learning exhibit significant violations of physical plausibility, including object collisions and object-out-of-room scenarios.

\begin{figure*}[t]
\centering
\includegraphics[width=\linewidth]{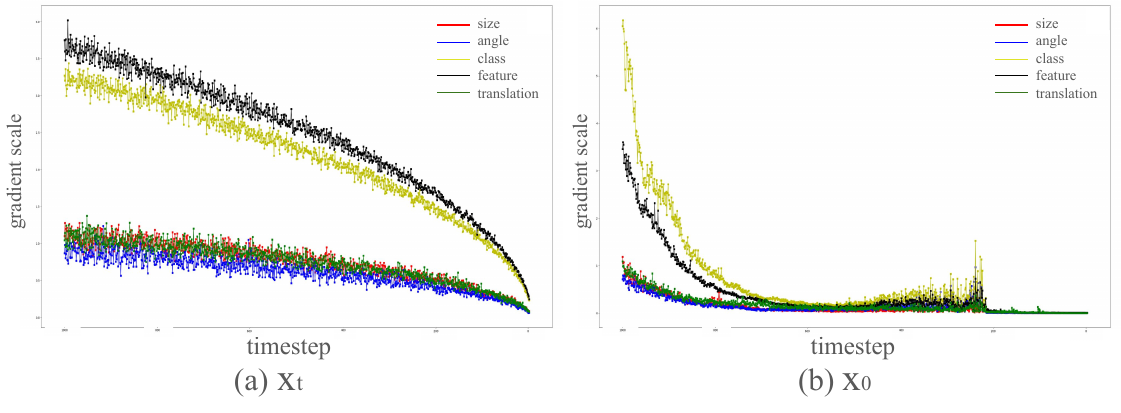}
\vspace{-15pt}
\caption{\textbf{Gradient scale varying with the denoising step.
}}
\label{fig:gradient_varying}
\end{figure*}

\begin{figure}[t!]
\centering
\includegraphics[width=1\linewidth]{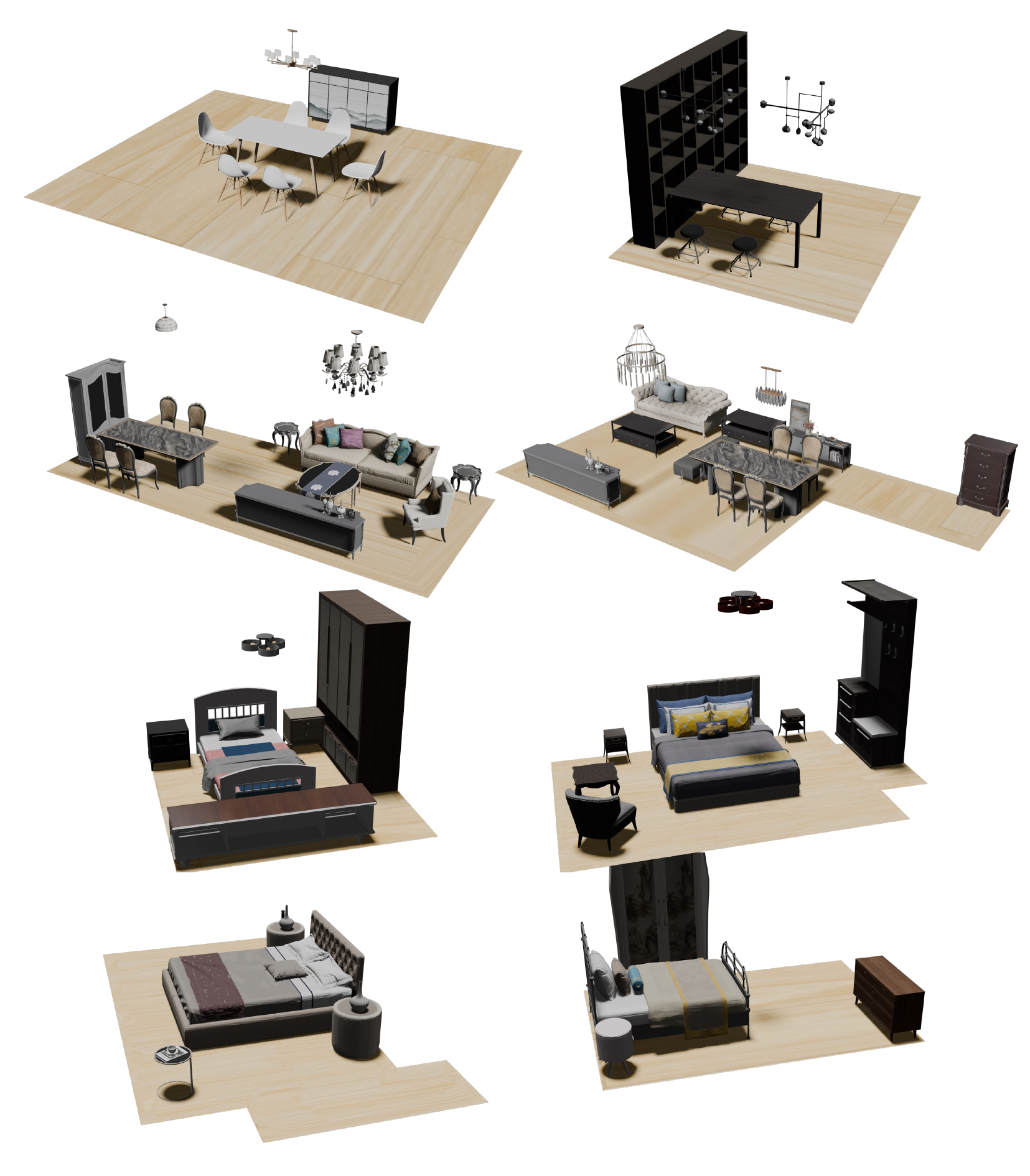}
\caption{\textbf{Visualization results of PhyScene on 3D Front.} The first two rows and the last two rows are the scene synthesis results of the Bedroom and Dining Room respectively.}
\label{fig:visual_beddining}
\end{figure}

\begin{table}[t!]
    \small
    \centering
    \caption{\textbf{Comparison against the original 3D-FRONT dataset on collision rate}. Both ATISS and DiffuScene have higher collision rates than the 3D-FRONT dataset, while ours is lower than 3D-FRONT in most cases. }
    \vspace{-10pt}
    \resizebox{\linewidth}{!}{
        \begin{tabular}{ccccccc}
            \toprule 
            \multirow{2}{*}{Data} & \multicolumn{2}{c}{Bedroom} & \multicolumn{2}{c}{Living Room} & \multicolumn{2}{c}{Dining Room}  \\
            \cmidrule(r){2-3} \cmidrule(r){4-5} \cmidrule(r){6-7} 
             & $\text{Col}_\text{obj}$ 
             & $\text{Col}_\text{scene}$ 
             & $\text{Col}_\text{obj}$ 
             & $\text{Col}_\text{scene}$ 
             & $\text{Col}_\text{obj}$ 
             & $\text{Col}_\text{scene}$ \\
             \midrule
            3D-FRONT  &  0.214  & 0.42 & 0.206 & \textbf{0.625} & 0.209 & 0.57  \\
            \midrule
            ATISS   &  0.248 & 0.46 & 0.316 &0.85 & 0.591 & 0.96 \\
            DiffuScene & 0.228  & 0.43 & 0.198 & 0.69 & 0.160 & 0.55 \\
            PhyScene(Ours)&  \textbf{0.187} & \textbf{0.36} & \textbf{0.191} & 0.63 & \textbf{0.151} & \textbf{0.53}  \\
            \bottomrule
        \end{tabular}
    }
    \label{tab:collision vs 3d front}
\end{table}


\subsection{Guidance on Different Agent Size}
\begin{figure}[t!]
\centering
\includegraphics[width=1\linewidth]{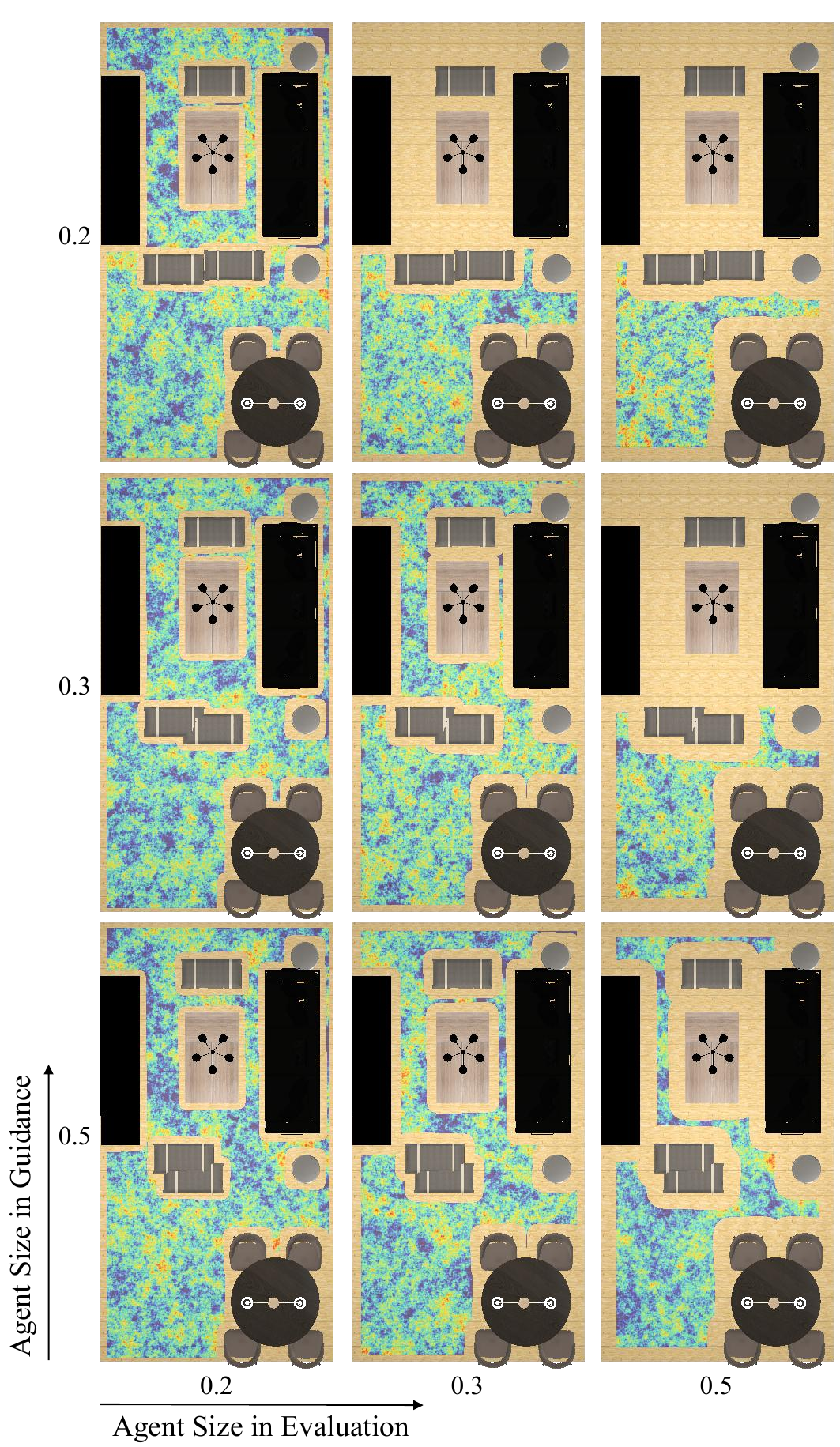}
\caption{\textbf{Reachability guidance results with different agent sizes.} We show the effectiveness of reachability guidance and the influence of the agent size. We compare walkable maps of different agent sizes both in guidance and in evaluation, which are 0.2, 0.3, and 0.5 separately. The unit of size is the meter. }
\label{fig:different_agent_size}
\end{figure}

The reachability guidance is adaptive to different agent sizes. We use 0.2, 0.3, and 0.5 as the agent size separately, where the unit of size is the meter. We show guidance results with different agent sizes in each row and evaluate each guided result on different agent sizes ( shown in each column). Here we show the guidance results in \cref{fig:different_agent_size} with the corresponding walkable map. It shows guidance on size 0.2 is not suitable for agent size 0.5, where the agent can only reach half of the room. And guidance on size 0.5 expands the walkable area to suit the agent in size of 0.5 and make the whole room reachable.

\section{Comparison with 3D FRONT}
Meanwhile, in \cref{tab:collision vs 3d front} we show models training on 3D-FRONT dataset can not get rid of the collision prior existed in the training dataset. Both ATISS and DiffuScene have higher collision rates on three types of rooms than 3D-FRONT. However, our PhyScene performs lower scores than 3D-FRONT. The result shows posterior optimization, such as physical and interactive guidance, is necessary to dismiss the unreasonable prior such as collision. 

\section{Guidance Details}
We visualize the gradient scale of each denoising step in \cref{fig:gradient_varying}. The gradient of $\bx_t$ decreases continuously during the denoising process, while the gradient of $\bx_0$ (predicted at each step) has a rapid decline at the beginning and intensively changes in the middle stage. We visualize the layout trajectory at each step and find the layout shrinks to the vicinity of the floor plan at the beginning stage and changes from chaos to order in the middle stage. The layout fine-tunes itself with slight changes at the final steps. According to this observation, we add guidance on the final steps. The results also confirm that adding guidance on the final steps performs the best.

When adding guidance to the data, our guidance is calculated by bounding box, including object size, location, and angle. The purpose is to make the layout more physically plausible and interactable. So we only calculate the gradient of location and angle for guidance to move objects into a more intractable position. Noting that guiding on size will lead to rather small sizes (thickness) of objects.

\begin{figure*}
    \centering
    \includegraphics[width=0.8\linewidth]{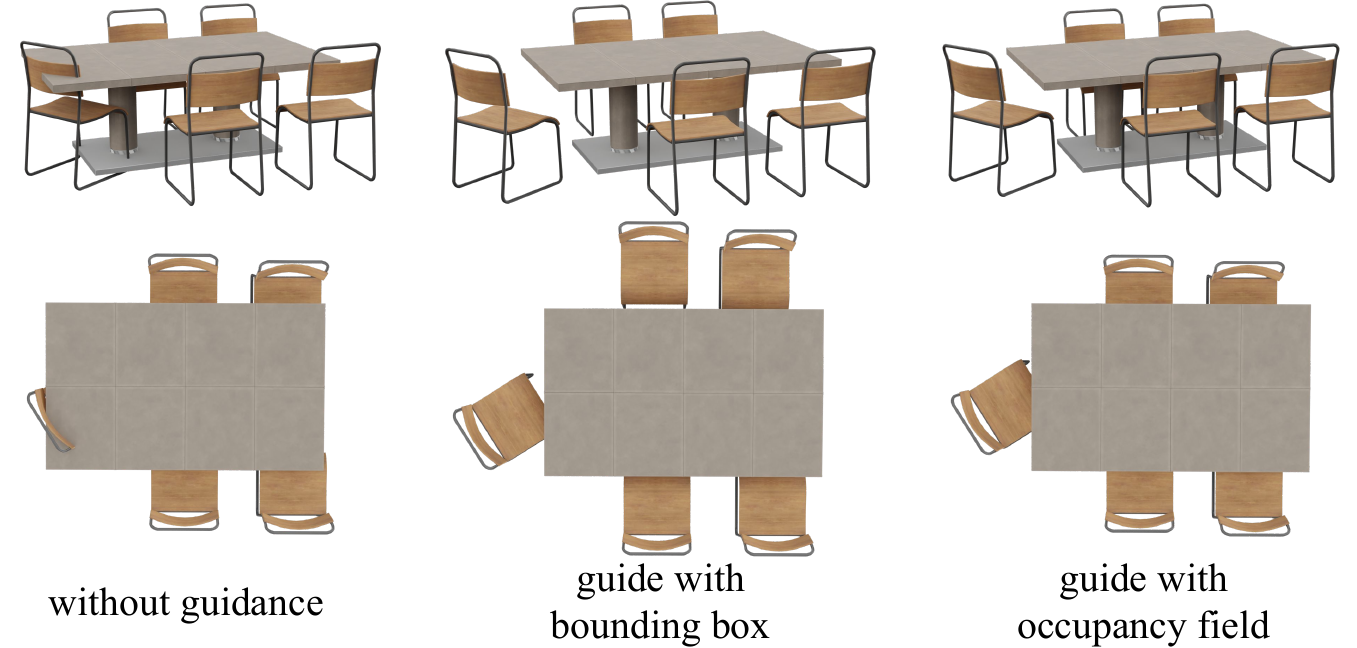}
    \caption{\textbf{Comparison of different 3D representations in collision guidance.} }
    \label{fig:sdf}
\end{figure*}
\section{Collision with Finer 3D Representations} 
In the collision guidance, we calculate the guidance objective on 3D bounding boxes of objects in \cref{eq:coll_guide}. We have also considered other finer representations (\eg, occupancy field). As the generation pipeline involves a non-differentiable object retrieval process from the generated object metadata (\ie, location, scale, \etc), using these finer 3D representations introduces non-trivial difficulty in model optimization. 
Nevertheless, we tried to use bounding boxes as representations for optimization while occupancy field collisions as indicators for loss calculation, \ie, using the following guidance function:

\begin{equation*}
\varphi_{\text{coll}}(\vx) = -\sum_{i,j, i\neq j} \textbf{IoU}_{3D}(\vb_i, \vb_j)\mathbbm{1}(\textbf{OF}(\vo_i,\vo_j)),
\end{equation*}

\noindent where $\mathbbm{1}(\textbf{OF}(\vo_i, \vo_j))$ checks if two objects have collided occupancy fields. This objective penalizes bounding box collisions only for objects that are collided in their corresponding occupancy fields. 

As shown in \cref{fig:sdf}, using occupancy fields as indicators can slightly improve the granularity of collision considered. However, as guidance calculation is required in multiple diffusion steps, computing the collision for two occupancy fields significantly increases the computation overhead (55 times slower). Therefore, we leave this exploration to find a better balance between speed and granularity using finer 3D representations as an important future work.









\section{Agent Interaction}
\label{sup:agent_interaction}
In the reachability guidance introduced in ~\cref{sub:reachability_guidance}, we only consider the walkable area as it is hard to unify guidance functions for object interactions, especially with various planners/modules required for different purposes (\eg, grasping, motion planning). However, as a preliminary attempt, we can extend the current pipeline to incorporate interaction constraints with proper simplifications.  
To ensure the articulated object interaction, we can use the same reachability guidance function while now 1) enlarging object bounding boxes to the maximum degree (fully opened) for recalculating the walkable map, 2) planning the shortest path from a walkable position to the end position of interactable object parts (\eg, drawer handles), and 3) applying the guidance to move the obstacle objects on this path. Similarly, we can model other interactions with rigid objects  (\eg, sit) by planning the shortest path to the interactive areas (\eg, space in front of the chair) correspondingly in the guidance function. 

With this simplified estimate, we can improve the \textit{interactiveness rate} (measured by whether robots could reach the end position of object parts when being maximum interacted) from 0.101 to 0.143. Given our flexible synthesize-with-guidance designs, we believe more fine-grained and effective constraints could be seamlessly integrated into the generation pipeline and will continue to explore this topic in the future.

\section{Diffusion v.s. Transformer}
ATISS uses an autoregressive model with an end vector to stop predicting new furniture, while we find the object number might be very large, such as predicting 33 objects in a bedroom.
In contrast, the diffusion model uses a fixed number of vectors and generates the objects' layout together. The predicted objects are embedded with overall information about the entire scene as well as inter-object relationships.

\section{Additional Visualization}
\begin{figure*}[t!]
\centering
\includegraphics[width=0.8\linewidth]{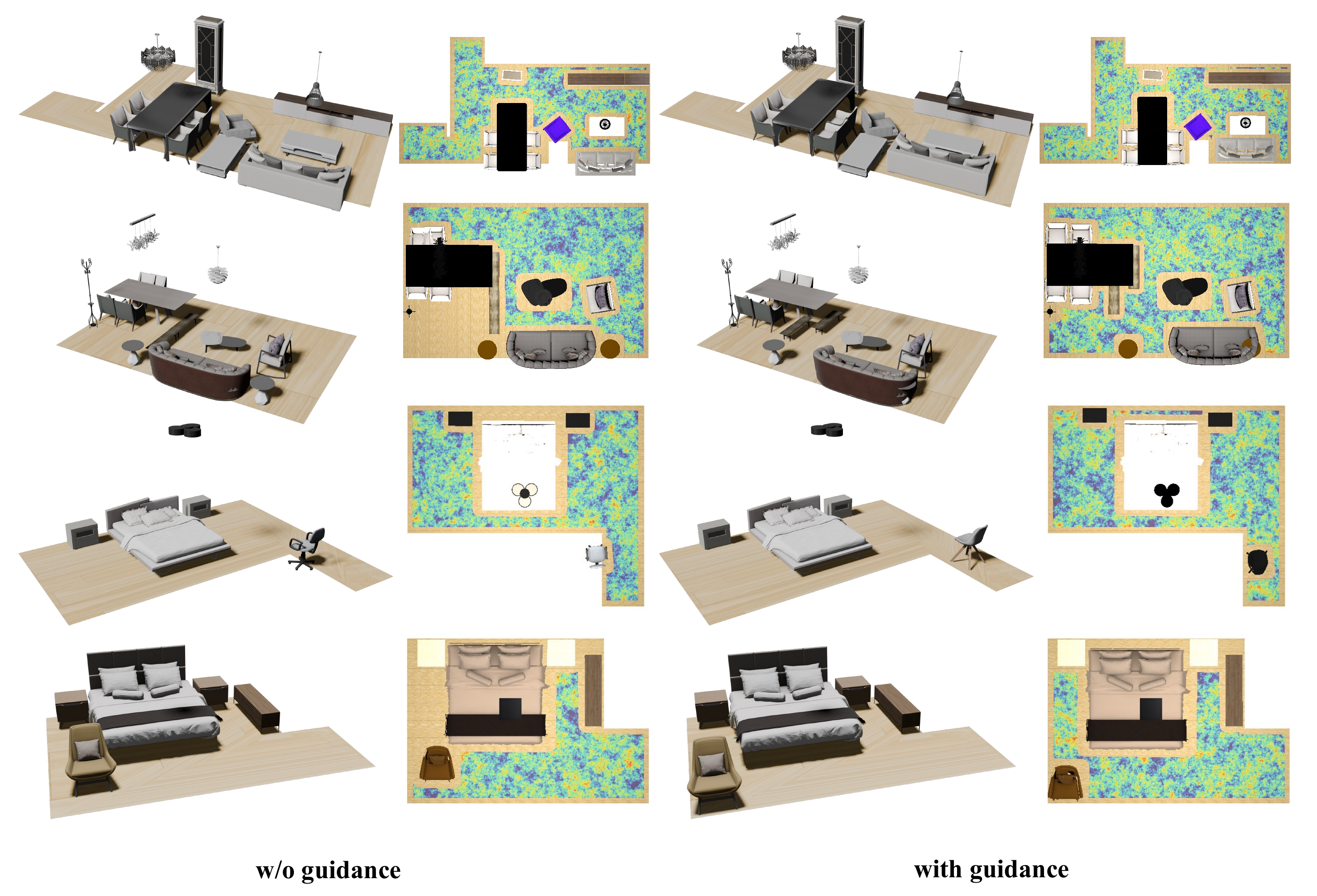}
\caption{\textbf{Comparison of PhyScene synthesis without and with guidance.}  The first two columns and the last two columns are the scene synthesis results without and with guidance respectively.}
\label{fig:beddiningroom}
\end{figure*}

\begin{figure*}[t!]
\centering
\includegraphics[width=0.8\linewidth]{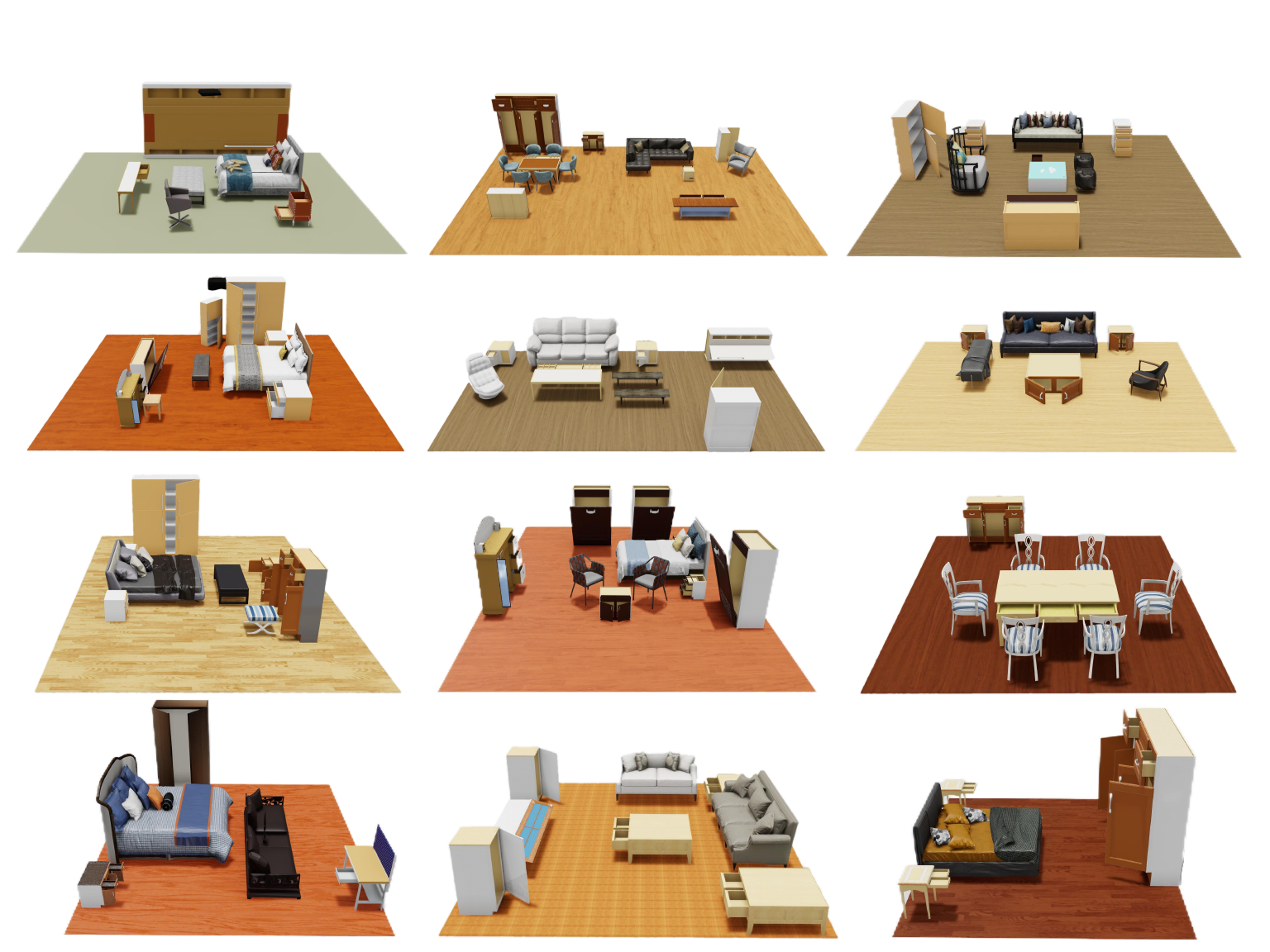}
\caption{\textbf{Generated scenes with articulated objects.} We show scene synthesis results with diverse layouts and random floor textures. Each scene is embedded with several articulated objects.}
\label{fig:procthor}
\end{figure*}

We provide additional qualitative visualization for the effectiveness of guidance functions in~\cref{fig:beddiningroom}. We also conduct experiments with basic floor plans (\ie, rectangles) in rooms from ProcTHOR and generate scenes with articulated objects. We provide the visualization of the generated results in~\cref{fig:procthor}.

%% file: camera_ready.bbl
\begin{thebibliography}{67}
\providecommand{\natexlab}[1]{#1}
\providecommand{\url}[1]{\texttt{#1}}
\expandafter\ifx\csname urlstyle\endcsname\relax
  \providecommand{\doi}[1]{doi: #1}\else
  \providecommand{\doi}{doi: \begingroup \urlstyle{rm}\Url}\fi

\bibitem[Ahn et~al.(2022)Ahn, Brohan, Brown, Chebotar, Cortes, David, Finn, Fu, Gopalakrishnan, Hausman, et~al.]{ahn2022can}
Michael Ahn, Anthony Brohan, Noah Brown, Yevgen Chebotar, Omar Cortes, Byron David, Chelsea Finn, Chuyuan Fu, Keerthana Gopalakrishnan, Karol Hausman, et~al.
\newblock Do as i can, not as i say: Grounding language in robotic affordances.
\newblock In \emph{Conference on Robot Learning (CoRL)}, 2022.

\bibitem[Anderson et~al.(2018)Anderson, Wu, Teney, Bruce, Johnson, S{\"u}nderhauf, Reid, Gould, and Van Den~Hengel]{anderson2018vision}
Peter Anderson, Qi Wu, Damien Teney, Jake Bruce, Mark Johnson, Niko S{\"u}nderhauf, Ian Reid, Stephen Gould, and Anton Van Den~Hengel.
\newblock Vision-and-language navigation: Interpreting visually-grounded navigation instructions in real environments.
\newblock In \emph{Proceedings of Conference on Computer Vision and Pattern Recognition (CVPR)}, 2018.

\bibitem[Bansal et~al.(2023)Bansal, Chu, Schwarzschild, Sengupta, Goldblum, Geiping, and Goldstein]{bansal2023universal}
Arpit Bansal, Hong-Min Chu, Avi Schwarzschild, Soumyadip Sengupta, Micah Goldblum, Jonas Geiping, and Tom Goldstein.
\newblock Universal guidance for diffusion models.
\newblock In \emph{Proceedings of Conference on Computer Vision and Pattern Recognition (CVPR)}, 2023.

\bibitem[Baruch et~al.(2021)Baruch, Chen, Dehghan, Dimry, Feigin, Fu, Gebauer, Joffe, Kurz, Schwartz, et~al.]{baruch2021arkitscenes}
Gilad Baruch, Zhuoyuan Chen, Afshin Dehghan, Tal Dimry, Yuri Feigin, Peter Fu, Thomas Gebauer, Brandon Joffe, Daniel Kurz, Arik Schwartz, et~al.
\newblock Arkitscenes: A diverse real-world dataset for 3d indoor scene understanding using mobile rgb-d data.
\newblock In \emph{Proceedings of Advances in Neural Information Processing Systems (NeurIPS)}, 2021.

\bibitem[Bi{\'n}kowski et~al.(2018)Bi{\'n}kowski, Sutherland, Arbel, and Gretton]{binkowski2018demystifying}
Miko{\l}aj Bi{\'n}kowski, Danica~J Sutherland, Michael Arbel, and Arthur Gretton.
\newblock Demystifying mmd gans.
\newblock \emph{arXiv preprint arXiv:1801.01401}, 2018.

\bibitem[Brooks et~al.(2023)Brooks, Holynski, and Efros]{brooks2023instructpix2pix}
Tim Brooks, Aleksander Holynski, and Alexei~A Efros.
\newblock Instructpix2pix: Learning to follow image editing instructions.
\newblock In \emph{Proceedings of Conference on Computer Vision and Pattern Recognition (CVPR)}, 2023.

\bibitem[Chang et~al.(2014)Chang, Savva, and Manning]{chang2014learning}
Angel Chang, Manolis Savva, and Christopher~D Manning.
\newblock Learning spatial knowledge for text to 3d scene generation.
\newblock In \emph{Proceedings of the conference on Empirical Methods in Natural Language Processing (EMNLP)}, 2014.

\bibitem[Dai et~al.(2017)Dai, Chang, Savva, Halber, Funkhouser, and Nie{\ss}ner]{dai2017scannet}
Angela Dai, Angel~X Chang, Manolis Savva, Maciej Halber, Thomas Funkhouser, and Matthias Nie{\ss}ner.
\newblock Scannet: Richly-annotated 3d reconstructions of indoor scenes.
\newblock In \emph{Proceedings of Conference on Computer Vision and Pattern Recognition (CVPR)}, 2017.

\bibitem[Deitke et~al.(2020)Deitke, Han, Herrasti, Kembhavi, Kolve, Mottaghi, Salvador, Schwenk, VanderBilt, Wallingford, et~al.]{deitke2020robothor}
Matt Deitke, Winson Han, Alvaro Herrasti, Aniruddha Kembhavi, Eric Kolve, Roozbeh Mottaghi, Jordi Salvador, Dustin Schwenk, Eli VanderBilt, Matthew Wallingford, et~al.
\newblock Robothor: An open simulation-to-real embodied ai platform.
\newblock In \emph{Proceedings of Conference on Computer Vision and Pattern Recognition (CVPR)}, 2020.

\bibitem[Deitke et~al.(2022)Deitke, VanderBilt, Herrasti, Weihs, Ehsani, Salvador, Han, Kolve, Kembhavi, and Mottaghi]{deitke2022procthor}
Matt Deitke, Eli VanderBilt, Alvaro Herrasti, Luca Weihs, Kiana Ehsani, Jordi Salvador, Winson Han, Eric Kolve, Aniruddha Kembhavi, and Roozbeh Mottaghi.
\newblock Procthor: Large-scale embodied ai using procedural generation.
\newblock In \emph{Proceedings of Advances in Neural Information Processing Systems (NeurIPS)}, 2022.

\bibitem[Dhamo et~al.(2021)Dhamo, Manhardt, Navab, and Tombari]{dhamo2021graph}
Helisa Dhamo, Fabian Manhardt, Nassir Navab, and Federico Tombari.
\newblock Graph-to-3d: End-to-end generation and manipulation of 3d scenes using scene graphs.
\newblock In \emph{Proceedings of International Conference on Computer Vision (ICCV)}, 2021.

\bibitem[Driess et~al.(2023)Driess, Xia, Sajjadi, Lynch, Chowdhery, Ichter, Wahid, Tompson, Vuong, Yu, et~al.]{driess2023palm}
Danny Driess, Fei Xia, Mehdi~SM Sajjadi, Corey Lynch, Aakanksha Chowdhery, Brian Ichter, Ayzaan Wahid, Jonathan Tompson, Quan Vuong, Tianhe Yu, et~al.
\newblock Palm-e: An embodied multimodal language model.
\newblock In \emph{Proceedings of International Conference on Machine Learning (ICML)}, 2023.

\bibitem[Fisher et~al.(2015)Fisher, Savva, Li, Hanrahan, and Nie{\ss}ner]{fisher2015activity}
Matthew Fisher, Manolis Savva, Yangyan Li, Pat Hanrahan, and Matthias Nie{\ss}ner.
\newblock Activity-centric scene synthesis for functional 3d scene modeling.
\newblock \emph{ACM Transactions on Graphics (TOG)}, 34\penalty0 (6):\penalty0 1--13, 2015.

\bibitem[Fu et~al.(2021{\natexlab{a}})Fu, Cai, Gao, Zhang, Wang, Li, Zeng, Sun, Jia, Zhao, et~al.]{fu20213d}
Huan Fu, Bowen Cai, Lin Gao, Ling-Xiao Zhang, Jiaming Wang, Cao Li, Qixun Zeng, Chengyue Sun, Rongfei Jia, Binqiang Zhao, et~al.
\newblock 3d-front: 3d furnished rooms with layouts and semantics.
\newblock In \emph{Proceedings of International Conference on Computer Vision (ICCV)}, 2021{\natexlab{a}}.

\bibitem[Fu et~al.(2021{\natexlab{b}})Fu, Jia, Gao, Gong, Zhao, Maybank, and Tao]{fu20213dfuture}
Huan Fu, Rongfei Jia, Lin Gao, Mingming Gong, Binqiang Zhao, Steve Maybank, and Dacheng Tao.
\newblock 3d-future: 3d furniture shape with texture.
\newblock \emph{International Journal of Computer Vision (IJCV)}, 129:\penalty0 3313--3337, 2021{\natexlab{b}}.

\bibitem[Fu et~al.(2017)Fu, Chen, Wang, Wen, Zhou, and Fu]{fu2017adaptive}
Qiang Fu, Xiaowu Chen, Xiaotian Wang, Sijia Wen, Bin Zhou, and Hongbo Fu.
\newblock Adaptive synthesis of indoor scenes via activity-associated object relation graphs.
\newblock \emph{ACM Transactions on Graphics (TOG)}, 36\penalty0 (6):\penalty0 1--13, 2017.

\bibitem[Geng et~al.(2023)Geng, Xu, Zhao, Xu, Yi, Huang, and Wang]{geng2023gapartnet}
Haoran Geng, Helin Xu, Chengyang Zhao, Chao Xu, Li Yi, Siyuan Huang, and He Wang.
\newblock Gapartnet: Cross-category domain-generalizable object perception and manipulation via generalizable and actionable parts.
\newblock In \emph{Proceedings of Conference on Computer Vision and Pattern Recognition (CVPR)}, 2023.

\bibitem[Gong et~al.(2023)Gong, Huang, Zhao, Geng, Gao, Wu, Ai, Zhou, Terzopoulos, Zhu, Jia, and Huang]{gong2023arnold}
Ran Gong, Jiangyong Huang, Yizhou Zhao, Haoran Geng, Xiaofeng Gao, Qingyang Wu, Wensi Ai, Ziheng Zhou, Demetri Terzopoulos, Song-Chun Zhu, Baoxiong Jia, and Siyuan Huang.
\newblock Arnold: A benchmark for language-grounded task learning with continuous states in realistic 3d scenes.
\newblock In \emph{Proceedings of International Conference on Computer Vision (ICCV)}, 2023.

\bibitem[Gu et~al.(2023)Gu, Xiang, Li, Ling, Liu, Mu, Tang, Tao, Wei, Yao, et~al.]{gu2023maniskill2}
Jiayuan Gu, Fanbo Xiang, Xuanlin Li, Zhan Ling, Xiqiang Liu, Tongzhou Mu, Yihe Tang, Stone Tao, Xinyue Wei, Yunchao Yao, et~al.
\newblock Maniskill2: A unified benchmark for generalizable manipulation skills.
\newblock \emph{arXiv preprint arXiv:2302.04659}, 2023.

\bibitem[Hart et~al.(1968)Hart, Nilsson, and Raphael]{Hart1968}
Peter Hart, Nils Nilsson, and Bertram Raphael.
\newblock A formal basis for the heuristic determination of minimum cost paths.
\newblock \emph{{IEEE} Transactions on Systems Science and Cybernetics}, 4\penalty0 (2):\penalty0 100--107, 1968.

\bibitem[Hassan et~al.(2019)Hassan, Choutas, Tzionas, and Black]{hassan2019resolving}
Mohamed Hassan, Vasileios Choutas, Dimitrios Tzionas, and Michael~J Black.
\newblock Resolving 3d human pose ambiguities with 3d scene constraints.
\newblock In \emph{Proceedings of International Conference on Computer Vision (ICCV)}, 2019.

\bibitem[Heusel et~al.(2017)Heusel, Ramsauer, Unterthiner, Nessler, and Hochreiter]{heusel2017gans}
Martin Heusel, Hubert Ramsauer, Thomas Unterthiner, Bernhard Nessler, and Sepp Hochreiter.
\newblock Gans trained by a two time-scale update rule converge to a local nash equilibrium.
\newblock In \emph{Proceedings of Advances in Neural Information Processing Systems (NeurIPS)}, 2017.

\bibitem[Ho and Salimans(2022)]{ho2022classifier}
Jonathan Ho and Tim Salimans.
\newblock Classifier-free diffusion guidance.
\newblock \emph{arXiv preprint arXiv:2207.12598}, 2022.

\bibitem[Ho et~al.(2020)Ho, Jain, and Abbeel]{ho2020denoising}
Jonathan Ho, Ajay Jain, and Pieter Abbeel.
\newblock Denoising diffusion probabilistic models.
\newblock In \emph{Proceedings of Advances in Neural Information Processing Systems (NeurIPS)}, 2020.

\bibitem[Huang et~al.(2023{\natexlab{a}})Huang, Yong, Ma, Linghu, Li, Wang, Li, Zhu, Jia, and Huang]{huang2023embodied}
Jiangyong Huang, Silong Yong, Xiaojian Ma, Xiongkun Linghu, Puhao Li, Yan Wang, Qing Li, Song-Chun Zhu, Baoxiong Jia, and Siyuan Huang.
\newblock An embodied generalist agent in 3d world.
\newblock \emph{arXiv preprint arXiv:2311.12871}, 2023{\natexlab{a}}.

\bibitem[Huang et~al.(2023{\natexlab{b}})Huang, Wang, Li, Jia, Liu, Zhu, Liang, and Zhu]{huang2023diffusion}
Siyuan Huang, Zan Wang, Puhao Li, Baoxiong Jia, Tengyu Liu, Yixin Zhu, Wei Liang, and Song-Chun Zhu.
\newblock Diffusion-based generation, optimization, and planning in 3d scenes.
\newblock In \emph{Proceedings of Conference on Computer Vision and Pattern Recognition (CVPR)}, 2023{\natexlab{b}}.

\bibitem[Huang et~al.(2022)Huang, Xia, Xiao, Chan, Liang, Florence, Zeng, Tompson, Mordatch, Chebotar, et~al.]{huang2022inner}
Wenlong Huang, Fei Xia, Ted Xiao, Harris Chan, Jacky Liang, Pete Florence, Andy Zeng, Jonathan Tompson, Igor Mordatch, Yevgen Chebotar, et~al.
\newblock Inner monologue: Embodied reasoning through planning with language models.
\newblock In \emph{Conference on Robot Learning (CoRL)}, 2022.

\bibitem[Inoue et~al.(2023)Inoue, Kikuchi, Simo-Serra, Otani, and Yamaguchi]{inoue2023layoutdm}
Naoto Inoue, Kotaro Kikuchi, Edgar Simo-Serra, Mayu Otani, and Kota Yamaguchi.
\newblock Layoutdm: Discrete diffusion model for controllable layout generation.
\newblock In \emph{Proceedings of Conference on Computer Vision and Pattern Recognition (CVPR)}, 2023.

\bibitem[Jia et~al.(2024)Jia, Chen, Yu, Wang, Niu, Liu, Li, and Huang]{jia2024sceneverse}
Baoxiong Jia, Yixin Chen, Huangyue Yu, Yan Wang, Xuesong Niu, Tengyu Liu, Qing Li, and Siyuan Huang.
\newblock Sceneverse: Scaling 3d vision-language learning for grounded scene understanding.
\newblock \emph{arXiv preprint arXiv:2401.09340}, 2024.

\bibitem[Jiang et~al.(2018)Jiang, Qi, Zhu, Huang, Lin, Yu, Terzopoulos, and Zhu]{jiang2018configurable}
Chenfanfu Jiang, Siyuan Qi, Yixin Zhu, Siyuan Huang, Jenny Lin, Lap-Fai Yu, Demetri Terzopoulos, and Song-Chun Zhu.
\newblock Configurable 3d scene synthesis and 2d image rendering with per-pixel ground truth using stochastic grammars.
\newblock \emph{International Journal of Computer Vision (IJCV)}, pages 920--941, 2018.

\bibitem[Jiang et~al.(2022)Jiang, Gupta, Zhang, Wang, Dou, Chen, Fei-Fei, Anandkumar, Zhu, and Fan]{jiang2022vima}
Yunfan Jiang, Agrim Gupta, Zichen Zhang, Guanzhi Wang, Yongqiang Dou, Yanjun Chen, Li Fei-Fei, Anima Anandkumar, Yuke Zhu, and Linxi Fan.
\newblock Vima: General robot manipulation with multimodal prompts.
\newblock \emph{arXiv}, 2022.

\bibitem[Khanna et~al.(2023)Khanna, Mao, Jiang, Haresh, Schacklett, Batra, Clegg, Undersander, Chang, and Savva]{khanna2023habitat}
Mukul Khanna, Yongsen Mao, Hanxiao Jiang, Sanjay Haresh, Brennan Schacklett, Dhruv Batra, Alexander Clegg, Eric Undersander, Angel~X Chang, and Manolis Savva.
\newblock Habitat synthetic scenes dataset (hssd-200): An analysis of 3d scene scale and realism tradeoffs for objectgoal navigation.
\newblock \emph{arXiv preprint arXiv:2306.11290}, 2023.

\bibitem[Kolve et~al.(2017)Kolve, Mottaghi, Han, VanderBilt, Weihs, Herrasti, Deitke, Ehsani, Gordon, Zhu, et~al.]{kolve2017ai2}
Eric Kolve, Roozbeh Mottaghi, Winson Han, Eli VanderBilt, Luca Weihs, Alvaro Herrasti, Matt Deitke, Kiana Ehsani, Daniel Gordon, Yuke Zhu, et~al.
\newblock Ai2-thor: An interactive 3d environment for visual ai.
\newblock \emph{arXiv preprint arXiv:1712.05474}, 2017.

\bibitem[Krantz et~al.(2020)Krantz, Wijmans, Majumdar, Batra, and Lee]{krantz2020beyond}
Jacob Krantz, Erik Wijmans, Arjun Majumdar, Dhruv Batra, and Stefan Lee.
\newblock Beyond the nav-graph: Vision-and-language navigation in continuous environments.
\newblock In \emph{Proceedings of European Conference on Computer Vision (ECCV)}, 2020.

\bibitem[Li et~al.(2021)Li, Xia, Mart{\'\i}n-Mart{\'\i}n, Lingelbach, Srivastava, Shen, Vainio, Gokmen, Dharan, Jain, et~al.]{li2021igibson}
Chengshu Li, Fei Xia, Roberto Mart{\'\i}n-Mart{\'\i}n, Michael Lingelbach, Sanjana Srivastava, Bokui Shen, Kent Vainio, Cem Gokmen, Gokul Dharan, Tanish Jain, et~al.
\newblock igibson 2.0: Object-centric simulation for robot learning of everyday household tasks.
\newblock \emph{arXiv preprint arXiv:2108.03272}, 2021.

\bibitem[Li et~al.(2023)Li, Zhang, Wong, Gokmen, Srivastava, Mart{\'\i}n-Mart{\'\i}n, Wang, Levine, Lingelbach, Sun, et~al.]{li2023behavior}
Chengshu Li, Ruohan Zhang, Josiah Wong, Cem Gokmen, Sanjana Srivastava, Roberto Mart{\'\i}n-Mart{\'\i}n, Chen Wang, Gabrael Levine, Michael Lingelbach, Jiankai Sun, et~al.
\newblock Behavior-1k: A benchmark for embodied ai with 1,000 everyday activities and realistic simulation.
\newblock In \emph{Conference on Robot Learning}, 2023.

\bibitem[Lin et~al.(2021)Lin, Wang, Olkin, and Held]{lin2021softgym}
Xingyu Lin, Yufei Wang, Jake Olkin, and David Held.
\newblock Softgym: Benchmarking deep reinforcement learning for deformable object manipulation.
\newblock In \emph{Conference on Robot Learning}, 2021.

\bibitem[Lu et~al.(2022)Lu, Zhou, Bao, Chen, Li, and Zhu]{lu2022dpm}
Cheng Lu, Yuhao Zhou, Fan Bao, Jianfei Chen, Chongxuan Li, and Jun Zhu.
\newblock Dpm-solver++: Fast solver for guided sampling of diffusion probabilistic models.
\newblock \emph{arXiv preprint arXiv:2211.01095}, 2022.

\bibitem[Mittal et~al.(2023)Mittal, Yu, Yu, Liu, Rudin, Hoeller, Yuan, Singh, Guo, Mazhar, et~al.]{mittal2023orbit}
Mayank Mittal, Calvin Yu, Qinxi Yu, Jingzhou Liu, Nikita Rudin, David Hoeller, Jia~Lin Yuan, Ritvik Singh, Yunrong Guo, Hammad Mazhar, et~al.
\newblock Orbit: A unified simulation framework for interactive robot learning environments.
\newblock \emph{IEEE Robotics and Automation Letters}, 2023.

\bibitem[Mu et~al.(2021)Mu, Ling, Xiang, Yang, Li, Tao, Huang, Jia, and Su]{mu2021maniskill}
Tongzhou Mu, Zhan Ling, Fanbo Xiang, Derek Yang, Xuanlin Li, Stone Tao, Zhiao Huang, Zhiwei Jia, and Hao Su.
\newblock Maniskill: Generalizable manipulation skill benchmark with large-scale demonstrations.
\newblock \emph{arXiv preprint arXiv:2107.14483}, 2021.

\bibitem[Nie et~al.(2023)Nie, Dai, Han, and Nie{\ss}ner]{nie2023learning}
Yinyu Nie, Angela Dai, Xiaoguang Han, and Matthias Nie{\ss}ner.
\newblock Learning 3d scene priors with 2d supervision.
\newblock In \emph{Proceedings of Conference on Computer Vision and Pattern Recognition (CVPR)}, 2023.

\bibitem[Paschalidou et~al.(2021)Paschalidou, Kar, Shugrina, Kreis, Geiger, and Fidler]{paschalidou2021atiss}
Despoina Paschalidou, Amlan Kar, Maria Shugrina, Karsten Kreis, Andreas Geiger, and Sanja Fidler.
\newblock Atiss: Autoregressive transformers for indoor scene synthesis.
\newblock In \emph{Proceedings of Advances in Neural Information Processing Systems (NeurIPS)}, 2021.

\bibitem[Poole et~al.(2022)Poole, Jain, Barron, and Mildenhall]{poole2022dreamfusion}
Ben Poole, Ajay Jain, Jonathan~T Barron, and Ben Mildenhall.
\newblock Dreamfusion: Text-to-3d using 2d diffusion.
\newblock \emph{arXiv preprint arXiv:2209.14988}, 2022.

\bibitem[Purkait et~al.(2020)Purkait, Zach, and Reid]{purkait2020sg}
Pulak Purkait, Christopher Zach, and Ian Reid.
\newblock Sg-vae: Scene grammar variational autoencoder to generate new indoor scenes.
\newblock In \emph{Proceedings of European Conference on Computer Vision (ECCV)}, 2020.

\bibitem[Qi et~al.(2018)Qi, Zhu, Huang, Jiang, and Zhu]{qi2018human}
Siyuan Qi, Yixin Zhu, Siyuan Huang, Chenfanfu Jiang, and Song-Chun Zhu.
\newblock Human-centric indoor scene synthesis using stochastic grammar.
\newblock In \emph{Proceedings of Conference on Computer Vision and Pattern Recognition (CVPR)}, 2018.

\bibitem[Ramesh et~al.(2022)Ramesh, Dhariwal, Nichol, Chu, and Chen]{ramesh2022hierarchical}
Aditya Ramesh, Prafulla Dhariwal, Alex Nichol, Casey Chu, and Mark Chen.
\newblock Hierarchical text-conditional image generation with clip latents.
\newblock \emph{arXiv preprint arXiv:2204.06125}, 1\penalty0 (2):\penalty0 3, 2022.

\bibitem[Ruan et~al.(2023)Ruan, Ma, Yang, He, Liu, Fu, Yuan, Jin, and Guo]{ruan2023mm}
Ludan Ruan, Yiyang Ma, Huan Yang, Huiguo He, Bei Liu, Jianlong Fu, Nicholas~Jing Yuan, Qin Jin, and Baining Guo.
\newblock Mm-diffusion: Learning multi-modal diffusion models for joint audio and video generation.
\newblock In \emph{Proceedings of Conference on Computer Vision and Pattern Recognition (CVPR)}, 2023.

\bibitem[Shridhar et~al.(2022)Shridhar, Manuelli, and Fox]{shridhar2022cliport}
Mohit Shridhar, Lucas Manuelli, and Dieter Fox.
\newblock Cliport: What and where pathways for robotic manipulation.
\newblock In \emph{Conference on Robot Learning}, 2022.

\bibitem[Song et~al.(2020)Song, Meng, and Ermon]{song2020denoising}
Jiaming Song, Chenlin Meng, and Stefano Ermon.
\newblock Denoising diffusion implicit models.
\newblock \emph{arXiv preprint arXiv:2010.02502}, 2020.

\bibitem[Szot et~al.(2021)Szot, Clegg, Undersander, Wijmans, Zhao, Turner, Maestre, Mukadam, Chaplot, Maksymets, et~al.]{szot2021habitat}
Andrew Szot, Alexander Clegg, Eric Undersander, Erik Wijmans, Yili Zhao, John Turner, Noah Maestre, Mustafa Mukadam, Devendra~Singh Chaplot, Oleksandr Maksymets, et~al.
\newblock Habitat 2.0: Training home assistants to rearrange their habitat.
\newblock \emph{Proceedings of Advances in Neural Information Processing Systems (NeurIPS)}, 34:\penalty0 251--266, 2021.

\bibitem[Tang et~al.(2024)Tang, Nie, Markhasin, Dai, Thies, and Nie{\ss}ner]{tang2023diffuscene}
Jiapeng Tang, Yinyu Nie, Lev Markhasin, Angela Dai, Justus Thies, and Matthias Nie{\ss}ner.
\newblock Diffuscene: Denoising diffusion models for gerative indoor scene synthesis.
\newblock In \emph{Proceedings of Conference on Computer Vision and Pattern Recognition (CVPR)}, 2024.

\bibitem[Wang et~al.(2021{\natexlab{a}})Wang, Xu, Xu, Liu, and Wang]{wang2021synthesizing}
Jiashun Wang, Huazhe Xu, Jingwei Xu, Sifei Liu, and Xiaolong Wang.
\newblock Synthesizing long-term 3d human motion and interaction in 3d scenes.
\newblock In \emph{Proceedings of Conference on Computer Vision and Pattern Recognition (CVPR)}, 2021{\natexlab{a}}.

\bibitem[Wang et~al.(2023)Wang, Zhao, Jiao, Zhu, Zhu, and Liu]{wang2023rearrange}
Weiqi Wang, Zihang Zhao, Ziyuan Jiao, Yixin Zhu, Song-Chun Zhu, and Hangxin Liu.
\newblock Rearrange indoor scenes for human-robot co-activity.
\newblock In \emph{Proceedings of International Conference on Robotics and Automation (ICRA)}, 2023.

\bibitem[Wang et~al.(2021{\natexlab{b}})Wang, Yeshwanth, and Nie{\ss}ner]{wang2021sceneformer}
Xinpeng Wang, Chandan Yeshwanth, and Matthias Nie{\ss}ner.
\newblock Sceneformer: Indoor scene generation with transformers.
\newblock In \emph{Proceedings of International Conference on 3D Vision (3DV)}, 2021{\natexlab{b}}.

\bibitem[Wang et~al.(2024)Wang, Chen, Jia, Li, Zhang, Zhang, Liu, Zhu, Liang, and Huang]{wang2024move}
Zan Wang, Yixin Chen, Baoxiong Jia, Puhao Li, Jinlu Zhang, Jingze Zhang, Tengyu Liu, Yixin Zhu, Wei Liang, and Siyuan Huang.
\newblock Move as you say, interact as you can: Language-guided human motion generation with scene affordance.
\newblock In \emph{Proceedings of Conference on Computer Vision and Pattern Recognition (CVPR)}, 2024.

\bibitem[Wu et~al.(2023)Wu, Ge, Wang, Lei, Gu, Shi, Hsu, Shan, Qie, and Shou]{wu2023tune}
Jay~Zhangjie Wu, Yixiao Ge, Xintao Wang, Stan~Weixian Lei, Yuchao Gu, Yufei Shi, Wynne Hsu, Ying Shan, Xiaohu Qie, and Mike~Zheng Shou.
\newblock Tune-a-video: One-shot tuning of image diffusion models for text-to-video generation.
\newblock In \emph{Proceedings of International Conference on Computer Vision (ICCV)}, 2023.

\bibitem[Xiang et~al.(2020)Xiang, Qin, Mo, Xia, Zhu, Liu, Liu, Jiang, Yuan, Wang, et~al.]{xiang2020sapien}
Fanbo Xiang, Yuzhe Qin, Kaichun Mo, Yikuan Xia, Hao Zhu, Fangchen Liu, Minghua Liu, Hanxiao Jiang, Yifu Yuan, He Wang, et~al.
\newblock Sapien: A simulated part-based interactive environment.
\newblock In \emph{Proceedings of Conference on Computer Vision and Pattern Recognition (CVPR)}, 2020.

\bibitem[Xu et~al.(2013)Xu, Chen, Fu, Sun, and Hu]{xu2013sketch2scene}
Kun Xu, Kang Chen, Hongbo Fu, Wei-Lun Sun, and Shi-Min Hu.
\newblock Sketch2scene: Sketch-based co-retrieval and co-placement of 3d models.
\newblock \emph{ACM Transactions on Graphics (TOG)}, 32\penalty0 (4):\penalty0 1--15, 2013.

\bibitem[Yang et~al.(2021)Yang, Guo, Zhou, and Tong]{yang2021indoor}
Ming-Jia Yang, Yu-Xiao Guo, Bin Zhou, and Xin Tong.
\newblock Indoor scene generation from a collection of semantic-segmented depth images.
\newblock In \emph{Proceedings of International Conference on Computer Vision (ICCV)}, 2021.

\bibitem[Yu et~al.(2022)Yu, Xie, Ma, Jia, Pang, Gao, Zhu, Zhu, and Wu]{yu2022latent}
Peiyu Yu, Sirui Xie, Xiaojian Ma, Baoxiong Jia, Bo Pang, Ruiqi Gao, Yixin Zhu, Song-Chun Zhu, and Ying~Nian Wu.
\newblock Latent diffusion energy-based model for interpretable text modeling.
\newblock In \emph{Proceedings of International Conference on Machine Learning (ICML)}, 2022.

\bibitem[Yuan et~al.(2023)Yuan, Song, Iqbal, Vahdat, and Kautz]{yuan2023physdiff}
Ye Yuan, Jiaming Song, Umar Iqbal, Arash Vahdat, and Jan Kautz.
\newblock Physdiff: Physics-guided human motion diffusion model.
\newblock In \emph{Proceedings of International Conference on Computer Vision (ICCV)}, 2023.

\bibitem[Zhai et~al.(2024)Zhai, {\"O}rnek, Wu, Di, Tombari, Navab, and Busam]{zhai2023commonscenes}
Guangyao Zhai, Evin~P{\i}nar {\"O}rnek, Shun-Cheng Wu, Yan Di, Federico Tombari, Nassir Navab, and Benjamin Busam.
\newblock Commonscenes: Generating commonsense 3d indoor scenes with scene graphs.
\newblock \emph{Proceedings of Advances in Neural Information Processing Systems (NeurIPS)}, 2024.

\bibitem[Zhang et~al.(2023)Zhang, Rao, and Agrawala]{zhang2023adding}
Lvmin Zhang, Anyi Rao, and Maneesh Agrawala.
\newblock Adding conditional control to text-to-image diffusion models.
\newblock In \emph{Proceedings of International Conference on Computer Vision (ICCV)}, 2023.

\bibitem[Zhang et~al.(2020)Zhang, Yang, Ma, Luo, Huth, Vouga, and Huang]{zhang2020deep}
Zaiwei Zhang, Zhenpei Yang, Chongyang Ma, Linjie Luo, Alexander Huth, Etienne Vouga, and Qixing Huang.
\newblock Deep generative modeling for scene synthesis via hybrid representations.
\newblock \emph{ACM Transactions on Graphics (TOG)}, 39\penalty0 (2):\penalty0 1--21, 2020.

\bibitem[Zheng et~al.(2022)Zheng, Chen, Jenkins, and Wang]{zheng2022vlmbench}
Kaizhi Zheng, Xiaotong Chen, Odest~Chadwicke Jenkins, and Xin Wang.
\newblock Vlmbench: A compositional benchmark for vision-and-language manipulation.
\newblock In \emph{Proceedings of Advances in Neural Information Processing Systems (NeurIPS)}, 2022.

\bibitem[Zhou et~al.(2019{\natexlab{a}})Zhou, Fang, Song, Guan, Yin, Dai, and Yang]{zhou2019iou}
Dingfu Zhou, Jin Fang, Xibin Song, Chenye Guan, Junbo Yin, Yuchao Dai, and Ruigang Yang.
\newblock Iou loss for 2d/3d object detection.
\newblock In \emph{Proceedings of International Conference on 3D Vision (3DV)}, 2019{\natexlab{a}}.

\bibitem[Zhou et~al.(2019{\natexlab{b}})Zhou, While, and Kalogerakis]{zhou2019scenegraphnet}
Yang Zhou, Zachary While, and Evangelos Kalogerakis.
\newblock Scenegraphnet: Neural message passing for 3d indoor scene augmentation.
\newblock In \emph{Proceedings of International Conference on Computer Vision (ICCV)}, 2019{\natexlab{b}}.

\end{thebibliography}
